\documentclass{article}
\usepackage{graphicx}
\usepackage[a4paper, total={6in, 8in}]{geometry}
\usepackage{amsfonts}
\usepackage{amssymb}
\usepackage{amsmath}
\usepackage{booktabs}
\usepackage{verbatim} 
\usepackage{markdown}
\usepackage{mdframed}

\usepackage{natbib} 

\usepackage{tabularx}   
\usepackage{tcolorbox}  
\usepackage{listings}  
\usepackage[T1]{fontenc}
\usepackage[utf8]{inputenc}

\newenvironment{practitionersummary}
  {\begin{center}
   \bfseries Practitioner Summary
   \end{center}
   \begin{quotation}\small}
  {\end{quotation}}

\usepackage{listings}
\usepackage{xcolor}
\usepackage{enumitem}

\usepackage{caption}
\usepackage[toc,page]{appendix} 

\usepackage{xcolor}
\usepackage{soul}

\usepackage{listings}
\usepackage{xcolor}

\lstdefinelanguage{json}{
  basicstyle=\ttfamily\footnotesize,
  showstringspaces=false,
  breaklines=true,
  morestring=[b]",
}

\lstdefinestyle{mypython}{
  language=Python,
  basicstyle=\ttfamily\footnotesize,
  numbers=left,
  numberstyle=\tiny,
  stepnumber=1,
  numbersep=5pt,
  frame=single,
  breaklines=true,
}

\usepackage{threeparttable}
\usepackage{hyperref}

\title{Generative Artificial Intelligence (GenAI) to convert images of queuing networks into verifiable simulation models: an open-weight LLM workflow approach}
\author{Thomas Monks, Alison Harper, Amy Heather, Navonil Mustafee}
\date{July 2026}

\begin{document}

\maketitle

\begin{abstract}
Recent work has explored the use of Large Language Models (LLMs) to automate simulation model building, typically by generating executable code directly from natural language descriptions. However, this raises challenges for verification and reproducibility particularly for users without programming expertise. We propose \textit{Sketch2DES}, a sketch-to-simulation workflow that converts diagrammatic representations of queuing networks into verifiable discrete-event simulation models using open-weight LLMs.
The workflow has three stages: (1) translation of a diagram into a semi-structured textual description using a multimodal LLM; (2) conversion into schema-validated structured data (JSON) via an LLM with a reflection-based verification loop; and (3) deterministic transformation into an executable simulation model using a software adapter. Intermediate artefacts can therefore be inspected and automatically validated before execution.
We evaluate the approach on eight queuing-network diagrams of varying complexity. The workflow achieved high reliability for all stages, and results were statistically indistinguishable from human-coded and analytical benchmarks.
Compared to direct code generation, the workflow improves reproducibility, transparency, and verifiability, while reducing the need for programming expertise. Limitations include restricted model scope and dependence on accurate visual interpretation. The results demonstrate the feasibility of structured, workflow-based model generation as a robust foundation for LLM-assisted simulation modelling.
\end{abstract}

\begin{practitionersummary}
Building simulation models typically requires specialist programming skills as well as expertise in conceptual modelling and verification. This can make simulation inaccessible to many domain experts, despite their understanding of the systems being modelled.

This study presents an AI workflow that converts sketches of queuing networks into executable discrete-event simulation models using Generative AI and software adaptors. Rather than generating simulation code directly, the workflow creates intermediate descriptions and structured model specifications that can be inspected and verified before a simulation model is produced. This allows users to identify and correct errors before implementation, increasing transparency and confidence in the resulting model.

For practitioners, the approach offers a practical way to reduce the effort required to translate conceptual models into executable simulations while maintaining human oversight. The workflow may be particularly valuable during early model development, stakeholder workshops, teaching, and rapid prototyping, where ideas are often communicated using informal diagrams rather than formal model specifications.

The current implementation is limited to a defined class of open queuing networks. However, the staged architecture provides a foundation for extending AI-assisted model generation to a broader range of simulation applications while maintaining reproducibility, transparency, and opportunities for human verification.
\end{practitionersummary}

\section{Introduction}

Previous work using Generative AI (GenAI) for Operational Research (OR) model building has shown promise \citep{Giabbanelli_GPT,jackson_2024,DEHGHANI_SIMGPT}, but has also exposed important methodological challenges, including model verification, cost, scalability, and privacy \citep{Zhou_GenAI_review_jors}. One approach to model generation is direct code generation \citep{jackson_2024}, in which a modeller prompts a Large Language Model (LLM), using natural language, to implement a simulation model in a programming language of their choice. However, studies have now shown that this approach can require substantial programming and modelling expertise from the user, particularly when generated code must be inspected, debugged, adapted, or verified \citep{Monks_Unlocking2025}. More fundamentally, direct generation of simulation code can produce materially different implementations across repeated runs, even when the same modelling problem is supplied. This variability raises concerns about reliability and trust, especially in settings where simulation models are used to inform operational decisions.

Research to date has predominantly relied on commercial frontier LLM services, including models such as Anthropic's Claude, Google's Gemini, and OpenAI's ChatGPT. Given the way access to LLMs has emerged, this is a logical choice for early studies. However, the proprietary nature of these GenAI tools means that the interfaces, behaviour, costs, licensing terms, and underlying models of these services can change during a study, or become unavailable at a later date. This makes reproducibility difficult and introduces the risk of model or version drift, where the performance of the LLM changes over time.

In this study, we introduce a three-stage LLM workflow approach to generating simulation models that sits within an emerging paradigm of sketch-to-simulation, or \textit{Sketch2Sim}; in particular we investigate Sketch to Discrete-Event Simulation (\textit{Sketch2DES}). Instead of asking an LLM to generate simulation code directly, we use open-weight LLMs, in a reproducible locally hosted workflow, for translation tasks: first from image to textual description, and then from description to structured data. The final conversion from structured data to a simulation model is deterministic, following a set of rules that we define. This approach is designed to mitigate the challenges associated with direct code generation and commercial AI services, including verification, reliability, privacy, reproducibility, and programming skill barriers. The workflow creates intermediate artefacts that can be inspected by a human modeller and allows automatic validation of the generated model specification before it is converted into executable simulation software. We demonstrate that the workflow can be executed privately using both academic High Performance Computing (HPC) infrastructure and a desktop machine equipped with a retail Graphics Processing Unit (GPU).

\section{Aims and Contributions}

This study investigates the feasibility of using open-weight LLMs and Free and Open-Source Software (FOSS) to convert diagrams of queuing networks into verifiable simulation models. Our motivation is both technical and practical. FOSS tools, in languages such as Python or R, offer transparency, reproducibility, and flexibility, but they often require programming skills that can create an entry barrier for novice modellers or modellers used to working with commercial off-the-shelf simulation software \citep{Smith2020ShinyModels,monks2023improving}. In contrast, many early-stage simulation studies begin with visual problem structuring: stakeholders and analysts sketch activities, resources, queues, routing decisions, and assumptions before any code is written. These process flow diagrams might be hand or computer drawn, use a variety of notations, and include different types of data. 

We therefore examine whether a workflow that starts from a diagram can reduce the need for users to directly implement models in code. The intention is not to remove the modeller from the process, but to support a workflow in which human effort is focused on defining, checking, and refining the conceptual and computer model, rather than on translating that model into the syntax and data structures required by a particular simulation package. 

This study makes three contributions. First, it introduces \textit{Sketch2DES}, a staged workflow that combines multimodal open-weight LLMs with inspectable intermediate artefacts and deterministic software transformation. Unlike direct code-generation approaches, the workflow enables errors to be identified separately during visual interpretation, formal model specification and software implementation. Second, it introduces a software-independent JSON representation for queuing-network DES models, together with an open-source \textit{json2ciw} adapter that automatically verifies the specification and converts it deterministically into an executable simulation model using the \textit{Ciw} package. This separates model specification from software implementation and provides a basis for extending to other simulation packages. Third, it provides a systematic evaluation of the workflow across repeated runs of eight queuing-network diagrams, quantifying stage-level and end-to-end reliability, computational effort, and failure modes, and identifying the points at which human-in-the-loop oversight remains necessary. 

Collectively, these contributions respond to recently identified research priorities for simulation. \cite{Boyle_grandchallenges} identify understanding how GenAI can support the construction and analysis of simulation models while providing valid and robust decision support as a grand challenge for the field (Grand Challenge 6), together with a more specific need for methods and tools that enable automated verification of AI-generated simulation models (Validation Challenge 4).  \textit{Sketch2DES} addresses these priorities by breaking model generation into steps that users can inspect and check before the final simulation model is built using reproducible, rule-based software.
\section{Related Literature}

\subsection{Automatic simulation model building} 

The challenge of building simulation models from human-readable specifications has motivated a sustained research effort over more than two decades. \citet{cimino2025automatic} provided a systematic review of 61 studies on Automatic Simulation Model Generation (ASMG) in industrial systems, identifying four broad development perspectives: data-centric approaches using event logs and process mining; algorithmic approaches using rule-based and machine learning driven model generation; knowledge-based approaches employing parametric templates, ontologies, and component libraries; and interface-facilitated approaches including graphical model builders. This tradition has generated a variety of flexible, reusable modelling frameworks, though manual intervention is required for all four strategies \citep{schmitt2026llm}.

A recurring theme across the literature is the use of structured intermediate representations to de-couple model specification from software-specific implementation. Using DES, efforts have included Business Process Model and Notation (BPMN) diagrams and similar formal notations \citep{onggo2011agent}. In the System Dynamics domain, the XMILE standard formalised this by defining an open XML interchange language that enables model sharing, reuse, and tool interoperability across platforms \citep{diker2005xmile}. While both traditions demonstrate that expressing a model in a software-agnostic structured format enables model generation and exchange, they require the modeller to create the formal specification manually before automation can proceed. Therefore, while structured representations improved interoperability and model generation, they did not remove the burden of formal model specification from the modeller.

The aspiration to use natural language rather than formal notation as the input for simulation model generation began in the 1980s. \citet{su1989natural} reported a working prototype that parsed English descriptions of manufacturing systems into a structured intermediate representation and automatically generated executable simulation code. This was explicitly motivated by the goal of software independence, but relied on a hand-crafted lexicon and case grammar rules constrained to a narrow manufacturing vocabulary. \citet{cimino2025automatic} noted that these challenges characterise the entire pre-LLM history of natural language interfaces for ASMG. The emergence of LLMs makes this approach feasible across a broader range of domains and input formats without requiring hand-crafted parsers, however they also introduce uncertainty, variability, and verification challenges.

\subsection{Multimodal and open-weight LLMs}

Recent advances in LLMs have extended model capabilities beyond text-only inputs. \textit{Multimodal} LLMs are models that can process and reason over multiple input types (modalities), such as text, images, tables, and diagrams within a unified architecture. This enables them to interpret relationships between visual and textual representations that may be included with a diagram of a queuing system, and generate structured outputs grounded in both.

Most existing simulation-generation frameworks take natural language descriptions as their primary input, either as written system descriptions \citep{Monks_Unlocking2025,jackson_2024}, or as iterative prompts \citep{Giabbanelli_GPT,DEHGHANI_SIMGPT}. \cite{giabbanelli2026text} examined the complementary use of text-to-image generation in modelling and simulation for communicating conceptual models, visualising simulation outputs, and generating educational materials. However, simulation studies commonly begin with visual problem structuring, sketching activities, queues, resources, and routing decisions on paper or whiteboards before any formal specification exists \citep{giabbanelli2024broadening,huang2025sketchgpt}. Multimodal LLMs can also process these visual artefacts alongside text, creating opportunities to treat diagrams as direct inputs to simulation model generation workflows.

Treating sketches as primary inputs to simulation pipelines predates LLMs. \citet{kara2004sim} demonstrated a working sketch-based interface for Simulink that converted hand-drawn block diagrams into functional simulation models using pen-stroke recognition and hierarchical symbol parsing, without any natural language component. More recently, \citet{ren2025simugen} revisited Simulink and demonstrated that multimodal LLMs could recreate models from the original Simulink diagram image when combined with a retrieval augmented generation (RAG) pipeline. Recent work has shown that multimodal LLMs can create formal model specifications directly from visual modelling artefacts. For example, \citet{bates2025unified} generated executable PlantUML representations from UML diagram images. However, these studies use diagrams in constrained or formalised formats (e.g. Simulink diagrams or UML), rather than informal, hand-drawn sketches of processes typical in early-stage DES modelling.

LLMs are typically accessed either through proprietary cloud-hosted services or through open-weight models that can be deployed locally. Proprietary \textit{frontier} LLMs have evidenced strong performance across a wide range of tasks, but their underlying weights, training data, and implementation details are generally unavailable to researchers. Additionally, their behaviour, version, and availability can be changed without notice, which creates challenges for scientific reproducibility. For example, proprietary LLMs such as GPT4 and o3 have been deprecated and superseded, making exact reproduction of any research using them impossible.

Open-weight models such as Meta’s LLaMA family or Google’s Gemma model partially address these concerns by making model weights publicly available under licenses that permit local deployment and redistribution. They can be downloaded, fine-tuned, and prompted on local hardware. This offers advantages for transparency, reproducibility, and data governance, particularly in domains where privacy requirements may restrict the external sharing of data \citep{riedemann2024openLLM}. Evidence suggests that open-weight models can achieve performance approaching that of proprietary frontier alternatives on a range of specialised tasks, including clinical information extraction \citep{open_weight_data_extraction,jabal2025open}, geospatial reasoning \citep{morandini2024correctness}, and more broadly across multilingual and multi-case structured extraction tasks \citep{spaanderman2025evaluating}. \citet{huang2025orlm} reported state-of-the-art results on optimisation modelling benchmarks using open-weight models. These developments suggest that open-weight LLMs are capable of supporting specialised model-generation tasks while offering advantages in reproducibility and governance for scientific simulation workflows.

\subsection{LLM workflows}

While LLMs provide powerful capabilities for information extraction, generation, and multimodal reasoning, their outputs remain probabilistic and may contain errors or inconsistencies. Consequently, increasing attention has been directed towards workflow-based approaches that separate generation, checking, and refinement tasks. Because verification and reliability are central concerns in simulation model generation, workflow architectures provide a promising mechanism for constraining and validating LLM outputs. Recent modelling and simulation guidance similarly emphasises task decomposition, explicit validation steps, and empirical evaluation of LLM behaviour and variability when designing multi-stage workflows \citep{giabbanelli2026guide}.

Workflows are structured pipelines in which one or more LLM calls are organised in a defined sequence, with the output of each call used as an input to a subsequent deterministic or LLM-driven step. Drawing on a taxonomy adopted by \citet{schmitt2026llm}, the most relevant workflow patterns for model generation are \textit{prompt chaining} (passing outputs between sequential LLM calls), the \textit{evaluator-optimiser loop} (where a generator produces an output and an evaluator critiques it and requests iterative revision), and \textit{parallelisation} (running multiple independent calls and selecting among the results). The evaluator-optimiser loop formalises a reflection pattern in which the generator is re-prompted with its own prior output and feedback to produce a corrected response. 

Related ideas have appeared in the literature, such as the Reflexion framework \citep{shinn2023reflexion} which demonstrated that iterative linguistic feedback can substantially improve performance on coding and reasoning tasks. Similar workflow principles have been applied in optimisation modelling, for example, \citet{Mostajabdaveh_Spec2Model_Opt} separated model generation from model verification through a multi-agent architecture, showing improved correctness and verifiability relative to direct generation approaches. In simulation modelling, \cite{schmitt2026llm} incorporated evaluator-optimiser loops and structured validation mechanisms to address hallucination and reliability challenges in automated model generation. Agentic systems represent a more flexible extension of workflow architectures by allowing LLMs to dynamically select tools and adapt execution paths. While such flexibility may be advantageous in open-ended tasks, workflow-based approaches offer stronger opportunities for deterministic validation and reproducibility when reliability is a primary concern. More broadly, workflow architectures enable complex generation tasks to be decomposed into smaller, inspectable artefacts that can be independently verified before subsequent stages are executed. This property is particularly attractive in simulation model generation, where correctness is often difficult to establish from executable code alone \citep{jackson_2024,schmitt2026llm}.

\subsection{GenAI for OR and simulation model generation}

The application of generative AI to OR has accelerated rapidly since the public release of capable LLMs. Recent reviews and editorial syntheses suggest that GenAI is becoming embedded across optimisation, stochastic systems, simulation, supply chains, and decision support, while simultaneously raising questions of verification, trustworthiness, and governance within OR workflows \citep{choi2026exploring,Zhou_GenAI_review_jors}. In simulation, \cite{Giabbanelli_GPT} provided an early systematic assessment of how LLMs could support simulation tasks, identifying model generation, parameter estimation, scenario design, and results identification as distinct areas of application, and establishing that LLM-generated prototype models could reduce the time required to reach a working implementation. Subsequently, \citet{giabbanelli2024broadening} examined the challenge of broadening simulation access to non-expert end-users via LLMs, identifying the gap between user intent and correctly specified model logic as the primary failure mode. This line of work was further extended to interoperability, proposing that LLMs can serve as middleware for converting model representations between heterogenous simulation platforms \citep{Giabbanelli_translators}. 

\citet{jackson_2024} presented a framework that translates natural language descriptions of logistics systems directly into executable DES models, demonstrating that an LLM could manage the full specification-to-code translation pipeline without formal intermediate notation. A related strategy in the healthcare context was implemented by \citet{Monks_Unlocking2025}, using GenAI to replicate simulation models from published descriptions of past studies. The authors demonstrated that LLMs can recover sufficient model structure from informal, narrative sources to produce valid implementations and reproduce published results. Extending this line of work, \citet{kute2025generative} proposed an automatic simulation model generation framework for factory planning that converts natural language descriptions into structured simulation data before instantiating executable models in Siemens Plant Simulation. Their approach incorporated validation and iterative optimisation loops, showing the potential value of intermediate machine-readable representations within LLM-supported simulation workflows. \citet{DEHGHANI_SIMGPT} introduced SimGPT, integrating LLMs with the Simio simulation environment across generation, execution, and analysis stages, and finding that LLM-based dispatching decisions can outperform traditional heuristics in job scheduling scenarios. The most comprehensive framework in manufacturing DES was developed by \citet{schmitt2026llm}, who proposed an end-to-end generative AI system for automated model generation, adaptation, and evaluation, explicitly addressing hallucination risk through evaluator-optimiser loops and structured output validation. 

A recurring concern is that generating simulation code directly from natural language descriptions produces outputs that are difficult to verify systematically. Unlike optimisation models, where feasibility and objective values provide useful correctness signals, simulation models often lack equivalent computable criteria, making verification of generated outputs particularly challenging. Similar challenges arise more broadly across OR applications that seek to transform unstructured inputs into formal analytical representations. For example, \cite{nie2026integrating} used a four-stage, auditable RAG pipeline to convert unstructured documentary evidence into validated, simulation-ready inputs for an agent-based model. Similarly, \citet{jackson2025supply} combined RAG with network science methods to extract supplier-customer relationships from public documents and automatically construct large-scale supply-chain graphs, showing how LLMs can act as intermediaries between unstructured information sources and structured OR artefacts. Recent simulation-generation frameworks have also begun to incorporate structured intermediate representations. For example, \citet{zhang2025intelligent} proposed a workflow in which product design documents are transformed into simulation models through an intermediate modelling language and scalable templates, reducing the complexity of direct code generation while enabling validation of generated artefacts.

The NL2OPT (natural language to optimisation model) literature has confronted a related challenge of translating informal problem descriptions into formally specified optimisation models. \citet{Ahmed_LM4Opt} introduced LM4OPT, a framework for progressively fine-tuning smaller open-weight LLMs for the NL2OPT problem, demonstrating that democratisation of access to optimisation modelling is achievable without reliance on large proprietary models. \citet{Mostajabdaveh_Spec2Model_Opt} addressed the verification problem more directly, proposing a multi-agent LLM framework that separates model generation from model verification, and providing evidence that structured specification inputs (Spec2Model) consistently outperform unstructured natural language descriptions (Desc2Model) in terms of both solution correctness and verifiability. \citet{huang2025orlm} extended this finding at scale in the OR domain, training open-weight models specifically for optimisation modelling, and achieving high performance on the LM4OPT benchmark. Structured intermediate representations reduce the verification problem from reasoning about arbitrary source code to checking a constrained set of model elements and relationships against predefined rules. Although optimisation differs from simulation, the NL2OPT literature directly addressed the problem of translating informal descriptions into formally checkable model specifications.

\subsection{Research gap}

Four strands of work are particularly relevant to this study: long-standing attempts to automate simulation model construction from human-readable inputs \citep{cimino2025automatic,su1989natural}; the rapid adoption of LLMs for this task across logistics, manufacturing, and healthcare simulation \citep{jackson_2024,schmitt2026llm,Monks_Unlocking2025}; the methodological case for structured intermediate representations as a means of improving reliability and enabling staged verification \citep{Mostajabdaveh_Spec2Model_Opt}; and the reproducibility and governance advantages of open-weight local deployment \citep{riedemann2024openLLM,manchanda2024open}. These developments suggest that simulation model generation is increasingly feasible, but also highlight unresolved challenges relating to verification, transparency, and reproducibility.

A specific methodological gap is apparent in the most directly comparable work of \citet{jackson_2024} and \citet{Monks_Unlocking2025}. These studies ultimately generate executable simulation code directly from natural language inputs without an intermediate representation that can be independently inspected and validated. \cite{schmitt2026llm} introduced additional validation and refinement stages, but the workflow remains centred on generating executable models rather than exposing a software-independent model specification that can be inspected and verified separately from the generated simulation code. While these approaches can produce working models, they introduce limitations. First, the absence of an inspectable intermediate representation means that verification is largely performed at the level of executable code rather than the conceptual model, limiting opportunities to identify structural errors before implementation. Second, reproducibility may be constrained by reliance on proprietary LLM services whose behaviour, availability, and underlying models may change over time. Finally, existing frameworks mostly assume natural language text as the primary input despite the fact that simulation studies often begin with sketches, process diagrams, BPMN-style process maps, and other visual artefacts produced during problem-structuring and process-mapping activities \citep{giabbanelli2024broadening}.

The final limitation is important because visual artefacts frequently constitute the earliest formal representation of a system and may never be translated into detailed textual specifications. Recent advances in multimodal LLMs make it possible to treat such diagrams as direct inputs to simulation model generation workflows. However, neither multimodal processing nor structured intermediate representations eliminate the need for conceptual validation. A schema-valid JSON specification or correctly generated simulation model can satisfy structural constraints while still failing to represent the intended real-world system. Schema validation can confirm that required fields are present, data types are correct, and constraints such as transition probabilities summing to one are satisfied, but it cannot establish conceptual truth. Consequently, the objective is not to automate conceptual validation, but to improve transparency and verifiability through the generation of inspectable intermediate artefacts that can be reviewed by a human before simulation is executed.

To our knowledge, no existing simulation-generation framework combines multimodal inputs, an inspectable intermediate specification, staged verification, and deployment using open-weight LLMs within a single workflow. The contribution of \textit{Sketch2DES} is the introduction of staged verification through inspectable intermediate artefacts, not the automation of conceptual validation. Rather than requiring verification to occur only after executable simulation code has been generated, the workflow enables errors to be identified at the levels of visual interpretation, model specification, and implementation.

\section{Sketch2DES workflow}

The workflow can be broken down into three stages:

\begin{enumerate}[
    label=\textbf{Stage~\arabic*:},
    leftmargin=6.0em
]
    \item Translate an image of a user's sketch of a queuing network into a semi-structured description;
    \item Translate the semi-structured text description of a model into verifiable structured data representing a model;
    \item Input structured data representing a model into a software adapter for a simulation package.
\end{enumerate}

The key contribution of the method is translation of unstructured data into structured data that can be inspected and verified.  This is very different from attempting to directly generate simulation logic in a language such as Python or R. 

\begin{figure}[htbp]
    \centering
    \includegraphics[width=\linewidth]{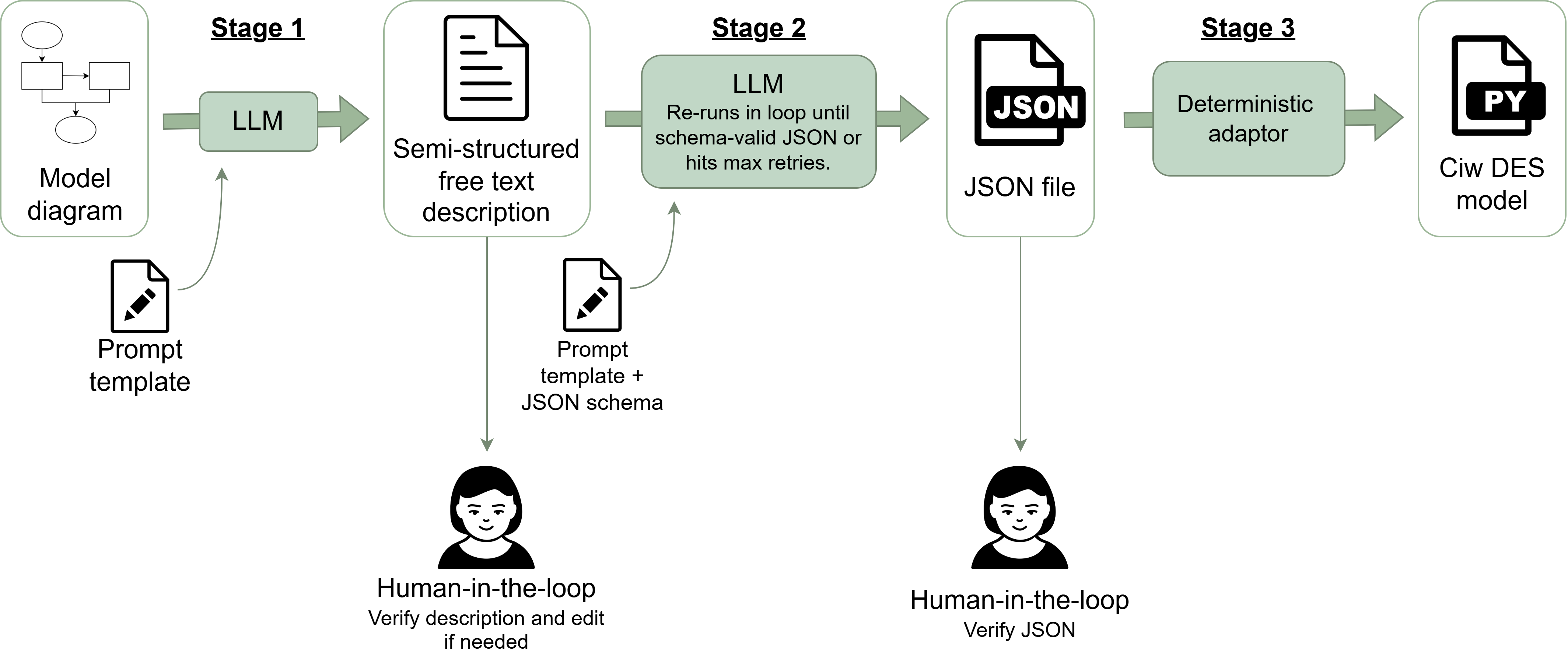}
    \caption{Overview of the three-stage Sketch2DES workflow implemented for the \textit{Ciw} DES package}
    \label{fig:method}
\end{figure}

A further advantage of our method is that it can be simplified or extended. For example, Stage 1 could be omitted if the description of the model already exists or has been generated in some other way, such as via ambient AI that summarises a problem structuring meeting where the system is described. In this study, executable models (Stage 3) are generated using \textit{ciw}, an  open-source Python library for queuing network simulation whose network representation aligns naturally with a JSON schema \citep{Palmer_ciw}. However, as the structured data we propose is software agnostic, it can be extended across simulation packages simply by creating new adapters. This means the method is generalisable across both commercial off-the-shelf tools and free and open source software. This approach shares a philosophy with the XMILE standard used across System Dynamics packages \citep{Martinez-Moyano_transparency}.  We describe the method in detail below and summarise the steps visually in Figure \ref{fig:method}. Following our description of the method, we illustrate its implementation with two applied examples in Section \ref{sec:applied_examples} and evaluate the method across eight examples in Section \ref{sec:evals}. 

\subsection{Stage 1: Semi-structured description generation}
The first stage of the workflow uses a multimodal LLM to generate a natural language description of a model from an input diagram. The LLM is provided with a prompt that defines the task, specifies diagram interpretation rules, and enforces a fixed output structure. The diagram image is supplied alongside this prompt. The model produces a description capturing key discrete-event simulation elements, including activities, resources, arrival processes, routing logic, and reneging behaviour.

The output is semi-structured: it remains human readable, while grouping related concepts (e.g., activities and routing) under fixed headings. This design supports human-in-the-loop verification, while enabling reliable parsing in the subsequent model extraction stage.

The prompt is tailored to DES and includes domain-specific rules, such as distinguishing activities from non-activity nodes, extracting parameterised distributions, and constructing routing probabilities. Table \ref{tab:stage1_prompt_summary} provides a summary of the prompt structure, with the full prompt included in Appendix \ref{app:stage1_prompt}.

The inclusion of domain rules increases consistency and accuracy (reliability) of the method. As an example, consider a healthcare analyst who has hand drawn a diagram of a hospital's emergency department (ED) queuing network. A challenge with LLMs is that they may contain general knowledge about EDs and could represent the ED in multiple varying ways. For example, an LLM may describe the ED in numbered process steps that a patient could follow, narrative prose and paragraphs, arbitrary groupings of concepts or patient groups, data formatted as markdown tables or bullet points, mermaid diagrams (a text-based standard for diagrams), or even programming language code. As we found when directly generating code to represent a DES model \citep{Monks_Unlocking2025}, each time the LLM is prompted to describe the diagram, the output format is likely to be different.  To increase the predictability of the LLM's response, our prompting approach defines a specialised LLM task, analysis requirements, and output format. In other words, each time a diagram is processed through Stage 1, the response is familiar and predictable.

\begin{table}[ht]
\centering
\caption{Summary of Stage 1 prompt sections, their purpose, and illustrative examples.}
\label{tab:stage1_prompt_summary}
\small
\begin{tabularx}{\linewidth}{p{3.2cm} p{4.5cm} p{5.5cm}}
\toprule
\textbf{Section} & \textbf{Purpose} & \textbf{Example} \\
\midrule
Role declaration &
  Primes the model as a DES and process flow expert &
  \emph{``You are an expert in analysing process flow diagrams\ldots''} \\
\addlinespace
Activity extraction (Rules 1, 7) &
  Identifies named process nodes and associated resources &
  Activity: \emph{Triage}; Resource: \emph{Nurse} (capacity=2) \\
\addlinespace
Distribution extraction (Rules 2, 5, 6) &
  Captures service, arrival, and reneging distributions with evaluated parameters &
  \texttt{exponential(mean=0.333)} converted from rate=3/hour \\
\addlinespace
Arrival logic (Rule 3) &
  Distinguishes external arrivals from internal routing &
  Poisson arrival at first queue vs.\ \emph{``none -- internal routing only''} \\
\addlinespace
Reneging/abandonment (Rule 4) &
  Extracts patience distributions only when explicitly shown in the diagram &
  \texttt{uniform(min=5, max=20)} minutes patience \\
\addlinespace
Node classification (Rules 9--12) &
  Prevents hallucinated activities by defining what is \emph{not} an activity &
  Exits, arrival nodes, reneging nodes, and timeless decision points excluded \\
\addlinespace
Routing matrix (Rules 13--14) &
  Builds a probabilistic transition table across activities and exits &
  After \emph{Triage}: Ward (0.7), EXIT (0.3); probabilities sum to 1.0 \\
\addlinespace
Ambiguity handling (Rule 15) &
  Requires explicit flagging rather than silent inference &
  \emph{``Service time illegible on Activity 3''} \\
\addlinespace
Output format &
  Provides a semi-structured response template &
  Fixed headings: \emph{Process Overview}, \emph{Activities}, \emph{Activity Routing}, \emph{Flow Logic} \\
\bottomrule
\end{tabularx}
\end{table}

\subsection{Stage 2: Structured data generation}

To address the verification and reliability challenges of models generated by AI tools, the core of our model building approach focuses on structured data. We chose JavaScript Object Notation (JSON), an open and human-readable standard for information exchange. The advantage of this approach is that structured data can be automatically verified against a pre-specified data schema. It is also agnostic to simulation software and programming languages.  Listing \ref{lst:processmodel-example} illustrates a simple urgent care process model in JSON. Appendix \ref{app:stage2_prompt} details the stage 2 prompt template, which can be applied across different JSON schemas and stage 1 model descriptions.

We aim to improve reliability through an LLM workflow called the reflection pattern. The LLM is provided with a system prompt specifying its purpose and task, a data schema, and a written description of the model (from stage 1). The generated structured data is then automatically tested for adherence to the schema. Example checks include whether the JSON is formatted correctly, whether JSON fields are of the appropriate types, and whether the field values are allowable. As an example of the latter, the verification step might check that valid distributions have been selected, and each distribution has a valid parameter, or in a queuing network all of the transition probabilities sum to one.  

If the verification fails, an informative verification error message is generated, and the reflection loop is triggered. The LLM is reprompted with a context including the original prompt, AI response, verification error and a request to fix. This loop is repeated until either the structured data passes verification or a maximum number of retries is reached. The latter condition might trigger a revision of the model description.

Common problems might include LLMs generating structured data inside formatted braces (e.g. ```) or prefixed with redundant conversational text such as ``here is your formatted JSON".  An optional step within the reflection cycle could therefore use deterministic text formatting rules to avoid expensive LLM calls.

The structured data is a digitally verifiable artefact representing the original sketch. The JSON can be stored and inspected by a human user, such as via a simple text editor. To aid human-in-the-loop verification, our approach also provides a visual representation of the JSON as a graph supplemented with extracted tables representing model parameters: distributions, resources, and routing matrices.

{\small
\begin{lstlisting}[language=json,
caption={ Example intermediate model artefact in JSON},
label={lst:processmodel-example}]
{
  "name": "Example urgent care process",
  "description": "Toy model with triage and treatment",
  "activities": [
    {
      "name": "Triage nurse",
      "type": "activity",
      "resource": {
        "name": "Nurse",
        "capacity": 2
      },
      "service_distribution": {
        "type": "exponential",
        "parameters": { "mean": 5.0 }
      },
      "arrival_distribution": {
        "type": "exponential",
        "parameters": { "mean": 3.0 }
      },
      "renege_distribution": null
    }
  ],
  "transitions": [
    {
      "from": "Triage nurse",
      "to": "Doctor consultation",
      "probability": 1.0
    }
  ]
}
\end{lstlisting}
}

\subsection{Stage 3: Structured data to simulation model}

The final stage of our LLM workflow does not require an LLM or any other form of GenAI. To increase the reliability of our method, we propose a deterministic approach where we define a mapping between our software-independent JSON format and the inputs required for a simulation package. We call this the adapter pattern. Each software package would require its own adapter; for example, \textit{SimPy} and \textit{Ciw} would have adapters in Python, \textit{Simmer} would have an adapter in R.  COTS packages, such as \textit{Simul8} and \textit{AnyLogic}, could similarly implement import routines that translate the structured representation into executable models, allowing users to open structured model specifications in much the same way as opening an existing model file.

To illustrate how an adapter layer works, consider the DES software \textit{Ciw}. An adapter layer must implement a deterministic mapping between JSON schema elements and \textit{Ciw} network constructs. Activities are translated into queuing nodes, resource capacities into server counts, arrival, service and reneging distributions into the corresponding \textit{Ciw} distribution objects, and transitions into a routing probability matrix. Because this transformation is rule-based rather than generative, identical JSON specifications always produce identical simulation models, ensuring that once a specification has been validated, model construction is fully reproducible. An adapter also performs deterministic consistency checks during model construction, for example validating that referenced nodes exist and that supported distribution types have been correctly specified. The current schema targets open queuing-network style DES models and supports arrivals, services, resources, routing, and reneging behaviour. Extension to other DES constructs such as schedules, batching, priorities, and state-dependent logic is left for future work.
\section{Applied examples}
\label{sec:applied_examples}

Before providing the results of a full evaluation of the method, we provide two end-to-end worked examples to illustrate the method in action, artefacts generated at each stage, and demonstrate the equivalence of key performance measures outputted by the LLM generated simulation model to one coded by a human. We emphasise that these results are intended to be explanatory and illustrative of the method.  Section \ref{sec:evals} provides a more detailed evaluation of each stage across different case studies.

\subsection{LLM, software and hardware environment}

We used open-weight LLMs on our own hardware, with no internet search capability or RAG, as part of our workflow using vLLM v0.19 for model inference and \textit{podman} v5.6.0 to containerise them.  All models were downloaded from \textit{Hugging Face}. We used Qwen3.5-35B-A3B (a mixture-of-experts architecture with 35B total parameters and 3B parameters active per token), at full precision, as a multimodal LLM and the smaller instruction-tuned Google's gemma-4-26b-A4B-it (a full precision, 26B parameters mixture of experts architecture with 4B active parameters) for structured data generation.  All steps were coded in Python 3.12. \textit{Langchain-openai} 1.1.12 was used to interact with Qwen3.5. We defined the JSON schema in Python using the \textit{pydantic} package v2.12.5.  This allowed both automatic verification of the generated JSON and the use of the \textit{instructor} v1.15.1 package to implement the reflection pattern with up to 4 retries.  In the evaluation, we set a random seed as part of each LLM call to allow our results to be reproduced. 

In Stages 1 and 2 LLMs process information as tokens, which are subword units representing fragments of text, words, punctuation, or other symbols. Both input and output sequence lengths are measured in tokens, making token counts a useful hardware independent proxy for computational effort, memory usage, inference cost, and latency. Throughout this study we therefore report input, reasoning, and output token usage for each stage of the workflow.

For multimodal inputs, an LLM first converts an image into a sequence of visual embeddings using a vision encoder. The visual embeddings are then projected into the same representational space as language tokens, allowing the model to process image and text information jointly within a single transformer architecture. Consequently, the token counts reported for Stage 1 include both textual prompt tokens and image-derived visual tokens generated from the input diagram.

For Stage 3, we wrote a small Python package called \textit{json2ciw} v0.10.0 \citep{monks_json2ciw} with functionality that adapts our JSON format to a DES model in the \textit{Ciw} v3.2.7 Python package. We chose \textit{Ciw} for this pilot as it has a permissive MIT open source license, good performance \citep{Palmer_ciw}, a clean usable interface based on a complex data structure, has automatic performance measure collection, and replications are simple to run across multiple CPUs. We designed \textit{json2ciw} to provide functionality to support human-in-the-loop verification of models by representing a JSON file as a Mermaid network diagram, and extract parameter, routing, and resource count tables.

The workflow and evaluation was executed on a Dell PowerEdge with an Intel Xeon Gold processor, 550GB RAM, NVIDIA H100 GPUs (a single H100 has 80GB of VRAM) running Red Hat Enterprise Linux (RHEL) 9.7. 

The code to recreate the two applied examples is documented in Jupyter notebooks \citep{monks_sketch2des_pilot}.

\subsection{Example 1: A hand-drawn call centre model}

For this example, we adapted the simple call centre model from \cite{monks2023improving}. The diagram of the model, Figure \ref{fig:applied_example_1}, was hand drawn by the authors.  

\subsubsection{Stage 1}

In Stage 1, the image was rotated by 90 degrees, converted to an in-memory PNG byte stream, and base64-encoded into a UTF-8 string. This encoded representation, together with a system prompt, was then provided to the LLM workflow to generate a semi-structured textual description of the process diagram.

\begin{figure}[htbp]
    \centering
    \begin{mdframed}
        \centering
        \includegraphics[width=0.5\textwidth, angle=90, origin=c]{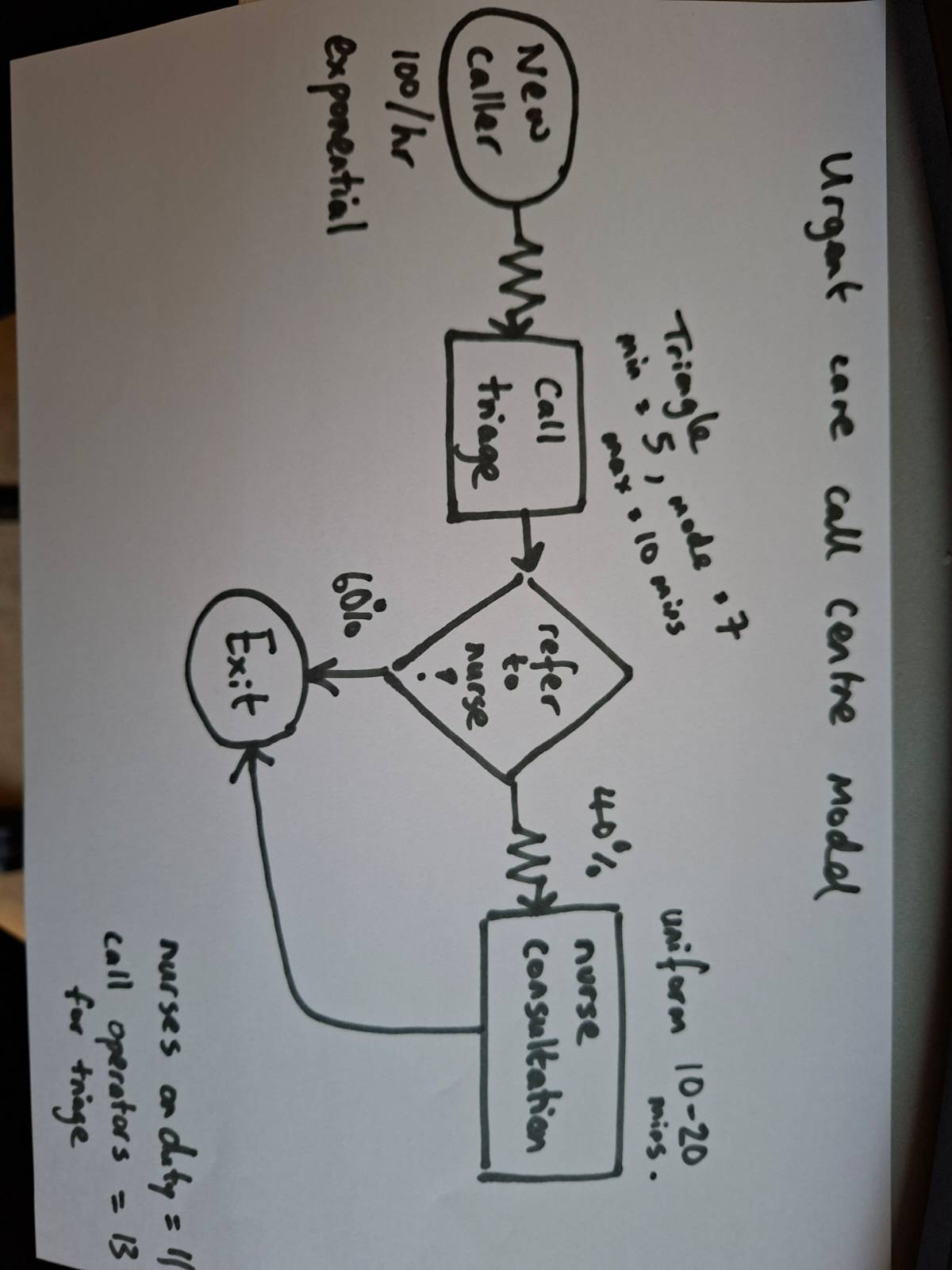}

        \vspace{0em}

        \begin{minipage}{0.9\textwidth}
        \footnotesize

        \textbf{Process Overview:}
        This diagram depicts a discrete event simulation model for an urgent care call center. The process begins with new callers arriving at the system. They first undergo a ``Call triage'' process handled by call operators. Following triage, a probabilistic routing decision determines whether a caller is referred to a nurse for consultation or exits the system immediately. If referred, they wait for and see a nurse before exiting.

        \textbf{Activities:}
        \begin{itemize}[leftmargin=*]
            \item \textbf{Activity 1: Call triage}
            \begin{itemize}[itemsep=0pt, topsep=1pt, parsep=0pt, partopsep=0pt]
                \item Service distribution: Triangle(min=5, mode=7, max=10) [units: mins]
                \item Arrival distribution: Exponential(rate=100/hr) [Converted to minutes: Exponential(mean=0.6 mins)]
                \item Reneging distribution: none
                \item Resource: call operators (capacity=13)
            \end{itemize}
            \item \textbf{Activity 2: nurse consultation}
            \begin{itemize}[itemsep=0pt, topsep=1pt, parsep=0pt, partopsep=0pt]
                \item Service distribution: Uniform(min=10, max=20) [units: mins]
                \item Arrival distribution: none -- internal routing only
                \item Reneging distribution: none
                \item Resource: nurses (capacity=11)
            \end{itemize}
        \end{itemize}

        \textbf{Activity Routing:}
        \begin{itemize}[leftmargin=*]
            \item Routing 1 (after Call triage):
            \begin{itemize}[itemsep=0pt, topsep=1pt, parsep=0pt, partopsep=0pt]
                \item Branch A (probability: 0.4) $\rightarrow$ nurse consultation
                \item Branch B (probability: 0.6) $\rightarrow$ EXIT
            \end{itemize}
            \item Routing 2 (after nurse consultation):
            \begin{itemize}[itemsep=0pt, topsep=1pt, parsep=0pt, partopsep=0pt]
                \item Branch A (probability: 1.0) $\rightarrow$ EXIT
            \end{itemize}
        \end{itemize}

        \textbf{Flow Logic:}
        \begin{enumerate}[leftmargin=*]
            \item \textbf{Arrival:} New callers enter the system with an inter-arrival time following an exponential distribution with a rate of 100 per hour (mean inter-arrival time of 0.6 minutes).
            \item \textbf{Triage:} Callers enter the ``Call triage'' queue (or are served immediately if resources are available). This activity is performed by 13 call operators. The service time is triangularly distributed between 5 and 10 minutes, with a most likely duration (mode) of 7 minutes.
            \item \textbf{Routing Decision:} After triage, callers reach a decision point.
            \begin{itemize}
                \item 40\% of callers are routed to ``nurse consultation''.
                \item 60\% of callers proceed directly to EXIT.
            \end{itemize}
            \item \textbf{Consultation:} The 40\% of callers referred to a nurse wait for one of the 11 available nurses. The service time for the consultation is uniformly distributed between 10 and 20 minutes.
            \item \textbf{Exit:} After the consultation (or after the 60\% routing branch), all callers leave the system.
        \end{enumerate}

        \end{minipage}
    \end{mdframed}
    \caption{Image provided to the workflow for example 1, with LLM-generated description.}
    \label{fig:applied_example_1}
\end{figure}

As shown in Figure \ref{fig:applied_example_1}, the model output is human-readable but semi-structured. Activities are defined individually, together with their associated resources, capacities, and arrival, service, and reneging distributions, while routing is described through explicit branching statements. The output also includes a higher-level textual overview of the process logic, making the overall flow of entities through the system easier to interpret.  In total stage 1 used 5,569 tokens: 2,769 inputted, 2,346 tokens used by Qwen3.5 for thinking, and 554 tokens to write the description.




\subsubsection{Stage 2}

In Stage 2 we combined our Stage 1 generated process description and the second prompt (see Appendix \ref{app:stage2_prompt}).  This was passed to the second LLM to generate a structured JSON file that is a specification of the simulation model to build.

Figure \ref{fig:json-mermaid-pair} illustrates the generated JSON file. It passed our automatic validation checks.  If it had failed, for example, due to an incorrect structure, the LLM would have been automatically re-prompted with the original prompt, output and details of the mistake to fix.  Figure \ref{fig:json-mermaid-pair} also illustrates the Mermaid diagram built automatically from the JSON. A total of 2,761 tokens were used in stage 2: 2,344 were input as the prompt, 0 were used for thinking, and 417 were output by Gemma4.

\subsubsection{Stage 3}

\begin{figure}[htbp]
    \centering
    \begin{minipage}[t]{0.48\linewidth}
        \vspace{0pt}
\begin{lstlisting}[
            language=json,
            basicstyle=\ttfamily\scriptsize,
            breaklines=true,
            frame=single,
            numbers=left
        ]
{
  "name": "Urgent Care Call Centre",
  "description": "Simulation model for an Urgent Care Call Centre.",
  "activities": [
    {
      "name": "Call triage",
      "type": "call_triage",
      "resource": {
        "name": "Call operators",
        "capacity": 13
      },
      "service_distribution": {
        "type": "triangular",
        "parameters": {
          "min": 5.0,
          "mode": 7.0,
          "max": 10.0
        }
      },
      "arrival_distribution": {
        "type": "exponential",
        "parameters": {
          "mean": 0.6
        }
      }
    },
    {
      "name": "Nurse consultation",
      "type": "nurse_consultation",
      "resource": {
        "name": "Nurses",
        "capacity": 11
      },
      "service_distribution": {
        "type": "uniform",
        "parameters": {
          "min": 10.0,
          "max": 20.0
        }
      },
      "arrival_distribution": null
    }
  ],
  "transitions": [
    {
      "from": "Call triage",
      "to": "Exit",
      "probability": 0.6
    },
    {
      "from": "Call triage",
      "to": "Nurse consultation",
      "probability": 0.4
    },
    {
      "from": "Nurse consultation",
      "to": "Exit",
      "probability": 1.0
    }
  ]
}
\end{lstlisting}
    \end{minipage}\hfill
    \begin{minipage}[t]{0.48\linewidth}
        \vspace{0pt}
        \centering
        \includegraphics[width=0.9\linewidth]{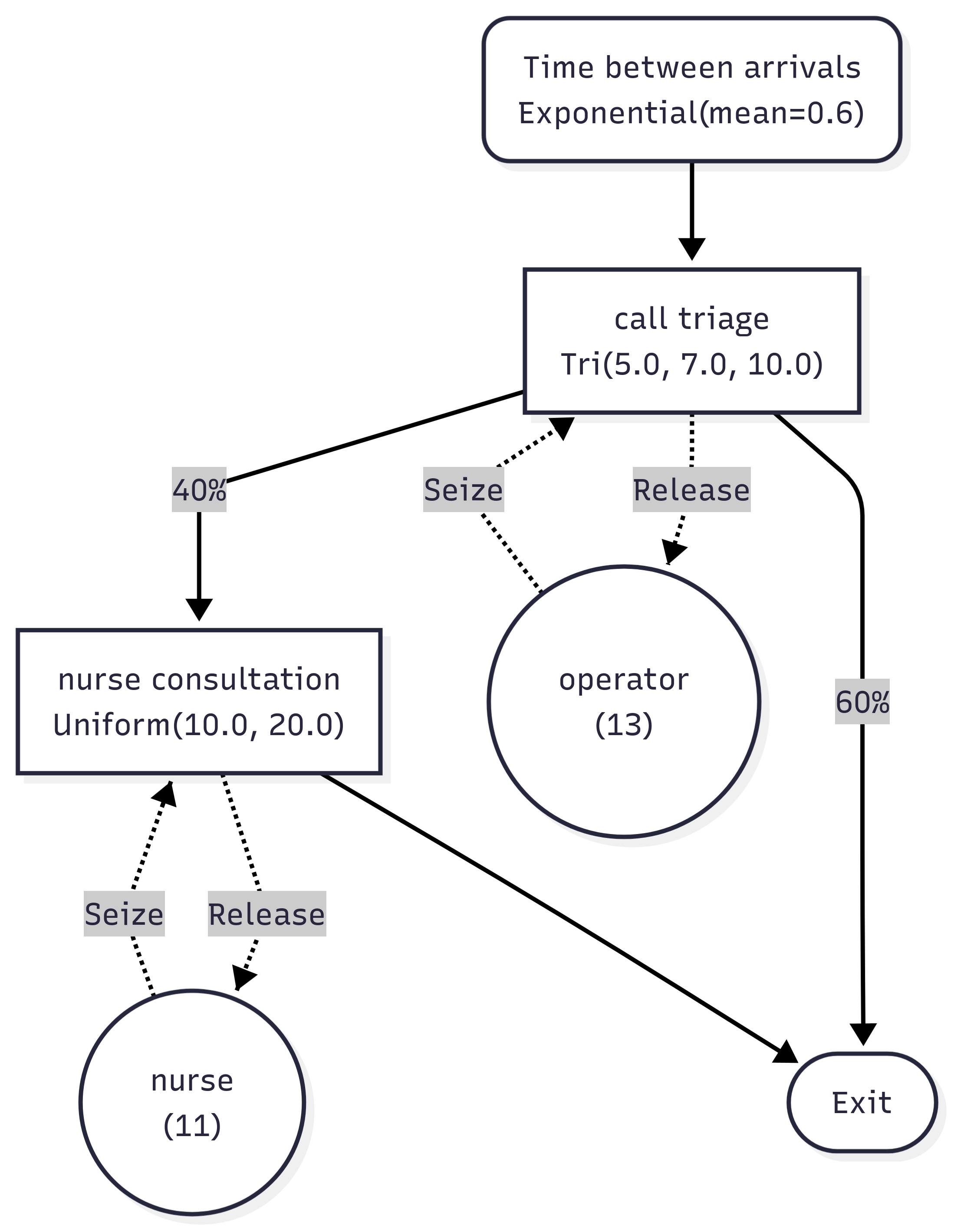}

        \vspace{0.75em}

        {\scriptsize \textbf{Ciw network model Python code}}\\[0.25em]
\begin{lstlisting}[
            style=mypython,
            language=Python,
            basicstyle=\ttfamily\scriptsize,
            breaklines=true,
            frame=single,
            numbers=none
        ]
import ciw

{
    'number_of_servers': [13, 11],
    'arrival_distributions': [
        ciw.dists.Exponential(rate=1.6666666666666667), 
        None
    ],
    'service_distributions': [
        ciw.dists.Triangular(lower=5.0, mode=7.0, upper=10.0), 
        ciw.dists.Uniform(lower=10.0, upper=20.0)
    ],
    'reneging_time_distributions': [None, None],
    'routing': [[0.0, 0.4], [0.0, 0.0]]
}
\end{lstlisting}
    \end{minipage}
    \caption{JSON model specification, Mermaid rendering, and final Ciw model generated in Stage 2 and 3.}
    \label{fig:json-mermaid-pair}
\end{figure}

The JSON representing the simulation model was deterministically converted into a Python data structure suitable for \textit{Ciw}. Figure \ref{fig:json-mermaid-pair} illustrates the data structure created in order to create and run a network model in \textit{Ciw}. As a validation step, in this example we followed the model comparison procedure recommended by \cite{sargent2020verification}, and the simultaneous confidence-interval procedure described by \cite{balci1994validation}. That is, we manually coded, tested and validated the call centre model using Python and \textit{SimPy} v4.1.1 and used this as a reference implementation for the model generated through our LLM workflow. We used \textit{SciPy} v1.17.1 functionality to construct Bonferroni-adjusted confidence intervals for the mean differences between the two implementations in waiting time, queue length and resource utilisation metrics.

Table \ref{tab:model_comparison} reports the statistics obtained from the AI-generated model in \textit{Ciw}, with the human coded and validated version of the model in \textit{SimPy}.  Each model had a warm-up period of 350 days, a run length of 10 days and 115 multiple replications. All of the Bonferroni-adjusted confidence intervals constructed contained zero, i.e., no statistically significant differences were detected between the AI generated model and our hand-coded SimPy model for any of the six performance measures.

\begin{table}[htbp]
\centering
\caption[Comparison of AI-generated and human-coded simulation models]{Comparison of mean performance measures between AI-generated and human-coded simulation models for example 1  ($\alpha = 0.05/6$).}
\label{tab:model_comparison}
\begin{tabular}{lrrrc}
\toprule
Simulation KPI & LLM & Human & $|\Delta|$ & 99.17\%$^*$ CI Diff \\
\midrule
Mean Waiting Time (operator) & 3.79 & 3.71 & 0.08 & (-0.20, 0.35) \\
Operator Utilisation & 94.01 & 93.93 & 0.08 & (-0.08, 0.24) \\
Operator Queue Length & 6.31 & 6.18 & 0.13 & (-0.35, 0.61) \\
Mean Waiting Time (nurse) & 4.48 & 4.18 & 0.30 & (-0.02, 0.63) \\
Nurse Utilisation & 90.90 & 90.89 & 0.02 & (-0.24, 0.27) \\
Nurse Queue Length & 2.99 & 2.79 & 0.20 & (-0.02, 0.43) \\
\bottomrule
\end{tabular}
\\\small$^*$95\% confidence intervals for the mean difference are Bonferroni corrected for 6 simultaneous comparisons $\alpha^* = 0.05/6$. A CI containing zero indicates no significant difference between models.
\end{table}

\subsection{Example 2: three node open Jackson Network}

We adapted an Open Jackson Network example from \cite{hillier2024introduction}. A Jackson Network consists of a set of servers (nodes) that can be modelled as set of connected M/M/s queues.  

\begin{figure}
    \centering
    \includegraphics[width=0.8\linewidth]{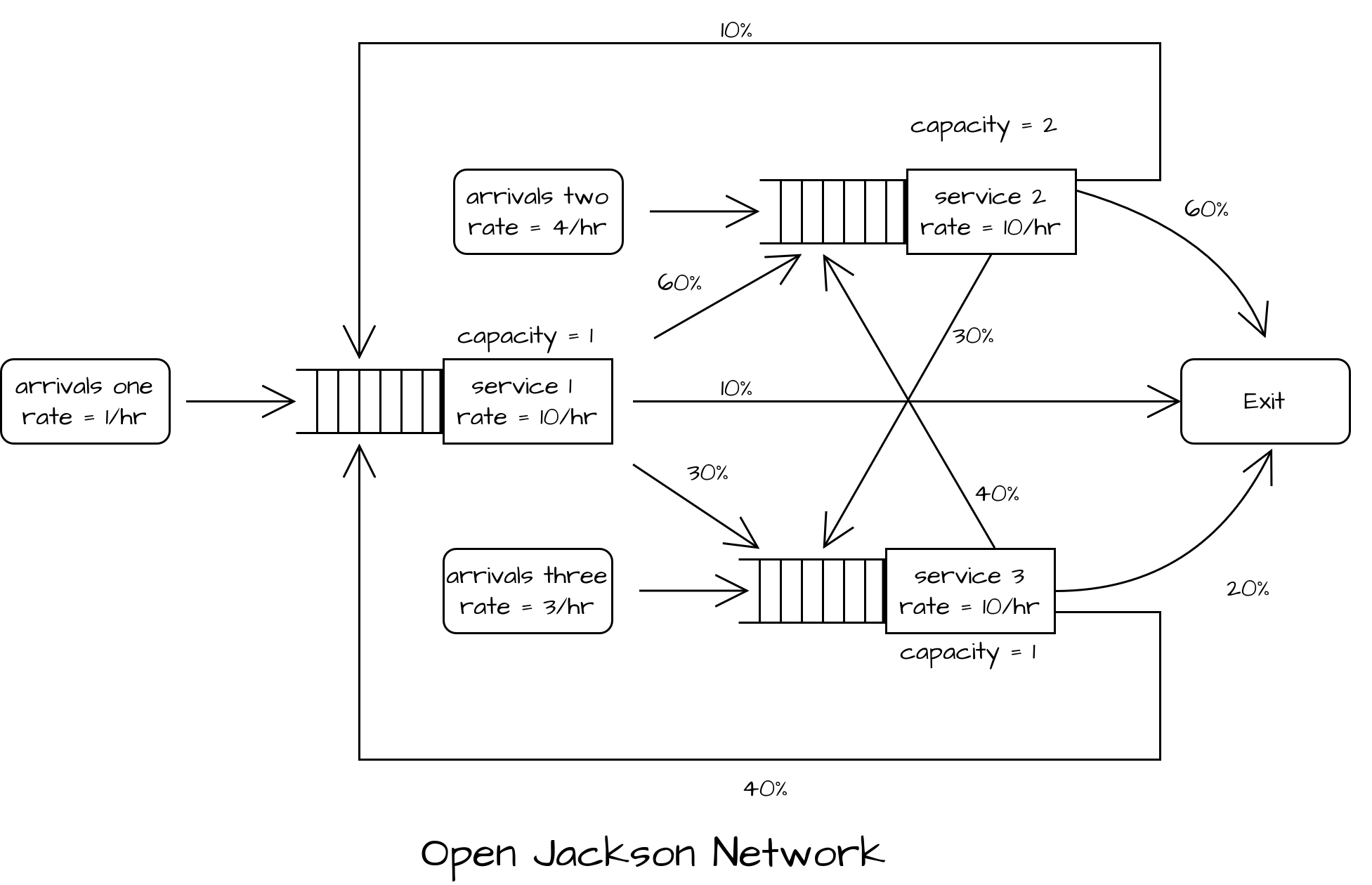}
    \caption{Example 2 image passed to the Stage 1 LLM}
    \label{fig:example_2_process_diagram}
\end{figure}

We created Figure \ref{fig:example_2_process_diagram}, a PNG image of the network in the software \textit{draw.io}, and passed it to the three-stage workflow as used in example 1.  To validate the LLM generated model in this example, we compared the simulation results to the exact analytical solutions using a two-sided one-sample t-test.  The results are presented in Table \ref{tab:llm_analytical_comparison}.  The AI-generated model had a warm-up period of 2880 time units, a run length of 14400 time units and 100 multiple replications. All of the Bonferroni-adjusted confidence intervals for the difference contained zero. This indicates that the performance metrics of the AI-generated simulation were statistically indistinguishable from the exact analytical solution of the Jackson Network.
Stage 1 used a total of 12,438 tokens (input=3,248, think=8,358, output=832) and stage 2 used a total of 3396 tokens (input=2,629, output=766).

\begin{table}[htbp]
\centering
\caption[Comparison of LLM-generated simulation results and analytical solution of the Jackson Network]{Comparison of LLM-generated simulation model and analytical queuing model across nine performance metrics ($W_q$ = mean waiting time; $\rho$ = server utilisation (\%); $L_q$ = mean queue length).}
\label{tab:llm_analytical_comparison}

\begin{tabular}{lrrrrc}
\toprule
 & LLM Generated & Analytical Model & $|\Delta|$ & $p$-value & 99.44\% CI Diff \\
Simulation KPI &  &  &  &  &  \\
\midrule
$W_q$ [service 1] & 0.10 & 0.10 & 0.00 & 0.2364 & (-0.0, 0.0) \\

$W_q$ [service 2] & 0.03 & 0.03 & 0.00 & 0.7299 & (-0.0, 0.0) \\
$W_q$ [service 3] & 0.30 & 0.30 & 0.00 & 0.4899 & (-0.0, 0.0) \\ \midrule
$\rho$ [service 1] & 49.98 & 50.00 & 0.02 & 0.5820 & (-0.1, 0.07) \\
$\rho$ [service 2] & 50.01 & 50.00 & 0.01 & 0.6324 & (-0.05, 0.07) \\
$\rho$ [service 3] & 75.01 & 75.00 & 0.01 & 0.8840 & (-0.12, 0.13) \\ \midrule
$L_q$ [service 1] & 0.50 & 0.50 & 0.00 & 0.3728 & (-0.0, 0.0) \\
$L_q$ [service 2] & 0.33 & 0.33 & 0.00 & 0.7713 & (-0.0, 0.0) \\
$L_q$ [service 3] & 2.26 & 2.25 & 0.01 & 0.4375 & (-0.02, 0.03) \\
\bottomrule
\end{tabular}
\begin{flushleft}
\footnotesize\textit{Note:} Bonferroni correction applied across 9 simultaneous comparisons: $\alpha^* = 0.05/9 \approx 0.006$. A CI for the mean difference containing zero indicates no significant difference between the LLM-generated simulation and the analytical model.
\end{flushleft}
\end{table}
\section{Stage-wise evaluation of the workflow}
\label{sec:evals}

To demonstrate the feasibility of our method, we implemented a pilot of the three-stage workflow for eight sketches of models, each of which were open queuing networks. Each node in a network could have its own stochastic arrival process, service distribution, and resource. To test an extension of the method beyond basic open networks we also allowed stochastic reneging (leaving queues before service). Entities leaving a node can probabilistically transition to one of $n$ linked nodes.  All transitions leaving a node must sum to one.  We used the same LLMs, plus software and hardware setup as described in the applied examples. Although human verification after Stages 1 and 2 is intrinsic to the workflow, the evaluation was conducted fully autonomously to assess the end-to-end performance of the automated pipeline.

\subsection{Overview of evaluation}

We conducted a stage-wise evaluation of the workflow that allowed us to analyse performance at each workflow stage before assessing end-to-end worflow. 

\begin{enumerate}
    \item Stage 1: Accuracy, reliability, and efficiency of the model description generation using LLM-as-a-judge classification, token tracking, and latency.
    \item Stage 2: Reliability and efficiency of generating the structured JSON and parallel \textit{Ciw} model from a \textit{valid} description of a queuing network.
    \item End-to-end workflow: reliability of generating a \textit{Ciw} model from an image without a human-in-the-loop.
\end{enumerate}

\subsection{Evaluation cases}

We created eight increasingly difficult open queuing network problems to test our workflow. Each of these was created either by the draw.io software or drawn by hand on paper (see appendix for all images). These were designed to test the workflow's ability to handle different numbers of network nodes (activities; i.e. 1-10), edges (routing), routing with loop backs to nodes, multiple arrival processes, multiple statistical distributions, parameters expressed with different time units and sources with different arrival rate notation (i.e., rates versus inter-arrival times), reneging, and different approaches to representing routing decisions in the process flow diagram. Our expectations were that hand-drawn diagrams, with more activities and routing, would pose the most challenge for the workflow.  For each case we created a ground truth description, i.e., the process flow diagram described in words, and simulation model in \textit{Ciw}. 

Table \ref{tab:eval_set} lists the eight case images in the evaluation set along with their characteristics. For example, case eight was hand drawn with 2 external arrivals sources, a 10 node network with 14 edges (connections between nodes) of which 2 were looping entities back to prior nodes, 2 queues with renege distributions, and all distributions were Exponential with parameters expressed in a single format.  In contrast case four, also hand drawn, had a single arrival source, 6 nodes, 11 edges with no looping back, 3 different distribution types (lognormal, exponential, and normal), with parameters expressed in one of 2 different time units (rate per day versus minutes).  Case 4 was adapted from \cite{nelson2013} and case 7 from \cite{gross1998fundamentals}.

\begin{table}[]
\centering
{\footnotesize 
\caption{The eight process flow diagrams used in the LLM workflow evaluation}
\label{tab:eval_set}
\begin{tabular}{@{}p{2.7cm}lllllp{1.0cm}p{0.8cm}l@{}}
\toprule
\textbf{Case}               & \textbf{Image} & \textbf{Sources} & \textbf{Nodes} & \textbf{Edges} & \textbf{Loops} & \textbf{Reneges} & \textbf{Dist Types} & \textbf{Units} \\ \midrule
1. Call Centre              & draw.io        & 1                & 2              & 4              & 0              & 0                & 3                      & 1              \\
2. Jackson Network          & draw.io        & 3                & 3              & 12             & 3              & 0                & 1                      & 1              \\
3. Call Centre 2            & hand           & 1                & 2              & 4              & 0              & 0                & 3                      & 2              \\
4. Treatment centre         & hand           & 1                & 6              & 11             & 0              & 0                & 3                      & 2              \\
5. MM1 with renege          & hand           & 1                & 1              & 2              & 0              & 1                & 1                      & 1              \\
6. Call centre renege & draw.io        & 1                & 2              & 4              & 0              & 1                & 3                      & 1              \\
7. Jackson 2        & hand           & 1                & 3              & 7              & 2              & 0                & 1                      & 3              \\
8. Ten node network         & hand           & 2                & 10             & 14             & 2              & 2                & 1                      & 1              \\ \bottomrule
\end{tabular}
}

\end{table}

\subsection{LLM-as-a-Judge}

The aim of Stage 1 evaluation is to assess whether the image‑to‑text stage produces faithful, usable model descriptions. We quantify this along three dimensions: correctness, reliability across repeated runs, and efficiency (latency and token usage).

As Stage 1 outputs are free‑form text and there are many acceptable ways to describe the same process, exact string matching is not suitable. We therefore adopt an LLM‑as‑a‑judge approach, using multiple independent LLMs of different families to assess whether a candidate description is faithful to a ground truth description of the queuing network. The judge’s task is to act as a binary classifier (\textit{correct} vs \textit{incorrect}) and to categorise any discrepancies into hallucinated activities, missing activities, misclassified nodes, missing distributions, parameter errors, routing errors, resource‑capacity errors, or LLM generation failure. A failed description could contain multiple discrepancies, e.g., two routing errors in one description, hence the error-classification counts we present for each case represent the number of identified errors rather than the number of descriptions that contained that type of error. See Appendix \ref{app:llm-judge} for the judge prompt and error schema.

We included three judges from different families. Two primary: OpenAI's gpt-oss-120b (117B parameters with 5B active parameters per token) plus Mistral-Small-3.2-24B-Instruct-2506  (a 24B parameter dense model, full precision); and one tiebreaker judge for when the primaries disagreed: Google's gemma-4-26b-A4B-it. We calibrated the judges by creating nine synthetic descriptions with known errors, i.e., human-modified ground truths to include different phrasing and mistakes. For these simple queuing network problems, the judges scored a perfect 100\% in binary classification, but there was some variance in the classification of the category of errors. For example, there was sometimes variance if a judge classified a node as misclassified or if it was hallucinated.  When applied to the case studies we randomly sampled 25 LLM judgements for human review. In each case the LLM-as-a-judge and human reviewer agreed.

\subsection{Evaluation metrics}

Generative AI is stochastic when temperature is greater than zero. This means that the same LLM will generate a different description given the same prompt.  For each evaluation case and stage, we therefore conducted multiple independent runs (Stage 1 = 50; Stage 2 = 20; end-to-end = 50). We held prompts constant across all experiments, controlled random seeds and used default temperatures across all models (Qwen3.5-35B-A3B = 1.0; gpt-oss-120b=1.0; gemma-4-26b-A4B-it=1.0; Mistral-Small-3.2-24B-Instruct-2506 = 0.15). We measure the repeated-sampling performance using $\mathrm{Pass@k}$ that estimates the probability that at least one of $k$ sampled outputs would be correct. In stage 1 this corresponds to the correct description, in stage 2 and the end-to-end evaluation this corresponds to the correct \textit{Ciw} model. \cite{chen2021evaluatinglargelanguagemodels} define $\mathrm{Pass@k}$ as:

\[
\widehat{\mathrm{Pass@k}} =  \mathbb{E}_{\text{problems}}\!\left[
  1 - \frac{\binom{n - c}{k}}{\binom{n}{k}}
\right]
\]

Here, \(n\) is the number of samples generated for the problem, \(c\) is the number classified as correct, and \(k\) is the number of candidate outputs considered.  In simple terms, you can interpret $\mathrm{Pass@1} = 0.80$ as a system with an 80\% chance of getting a valid description of your queuing network on the first attempt. $\mathrm{Pass@2}=1.00$ means that, when two outputs are sampled, the estimated probability that at least one of them is correct is 100\%.  We report $\mathrm{Pass@k}$ for each case, and overall (the mean of the case-level estimates) for $k$ = 1, and 2.

Our final two metrics focussed on efficiency. Latency, measured in seconds, was calculated from prompt submission to completion of the final token and excluded image preprocessing. Token usage was collected for input (image + prompt template), model reasoning (Qwen3.5 uses \textit{thinking} tokens before generating an answer) and output (the response shown to the user). For multimodal inputs, image content contributes substantially to token usage because the vision encoder represents image regions as visual tokens in addition to the textual prompt and generated response.

\subsection{Evaluation Results}

Overall, the workflow converted images in our evaluation set of queuing networks into expected simulation models and intermediate JSON artefacts with high reliability. Tables \ref{tab:stage1_baseline_compact}, \ref{tab:stage2_results} and \ref{tab:end2end_results} break the results down for Stage 1, 2, and the end-to-end workflow respectively.

For Stage 1 the observed Pass@1 was 0.97 (389/400 descriptions judged as correct; range = 0.9-1.0). If we selected 2 of these runs at random then it was very certain that one would be a fully accurate description (Pass@2 = 1.0; range = 0.99-1.0).  The most likely failure mode was incorrect routing between nodes. Case 8, with 10 nodes, reneging, and routing loop backs, had the most failures (5/50) and also required the most tokens (mean=22461, stdev=2270). On our hardware it took an average of 30 seconds (stdev=9s, range 17-54s) to generate a description.

In Stage 2 we tested two models: Gemma 4 and Mistral Small 3.2. Accuracy and reliability of the models was identical: in each case tested the method returned the expected \textit{Ciw} model (Pass@1 = 1.0). The workflow was fast for both models ranging between 4 seconds and 17 seconds to generate a valid JSON structure representing the simulation model, but notably quicker for Gemma 4 (mean = 5s) compared to Mistral Small 3.2 (mean = 10s). Total token usage was lower than Stage 1 with average usage ranging from 2733 to 4404 across the eight cases.

The end-to-end workflow successfully generated the expected \textit{Ciw} models in 388 of the 400 runs across the eight cases (Pass@1 = 0.97 and Pass@2 = 1.0). There were no instances where the workflow failed to complete.  On average time to complete the workflow was 35 seconds (std = 16s) and ranged from 21 to 65 seconds across the eight cases. The total number of tokens used averaged 16942 (range 9053-27514).

\begin{table}
\centering
{\footnotesize
\setlength{\tabcolsep}{3pt}
\caption{Performance of Stage 1 image to text; overall and across all evaluation cases.}
\label{tab:stage1_baseline_compact}
\begin{tabular}{l l r r r r r r r r r r r r r}
\toprule
& & & & & \multicolumn{8}{c}{Error Classification} & \\
\cmidrule(lr){6-13}
Case & Pass@1 & Pass@2 & Latency (s) & Tokens & Hallu. & Misclass. & MisNode & MisDist & Param. & Route & Res & Gen. & Passes \\
\midrule
1 & 1.00 & 1.00 & 17.6 & 6251.5 & 0 & 0 & 0 & 0 & 0 & 0 & 0 & 0 & 50/50 \\
2 & 1.00 & 1.00 & 32.1 & 9326.6 & 0 & 0 & 0 & 0 & 0 & 0 & 0 & 0 & 50/50 \\
3 & 1.00 & 1.00 & 16.5 & 5859.4 & 0 & 0 & 0 & 0 & 0 & 0 & 0 & 0 & 50/50 \\
4 & 0.96 & 1.00 & 34.0 & 18815.7 & 0 & 2 & 0 & 0 & 0 & 0 & 0 & 0 & 48/50 \\
5 & 0.96 & 1.00 & 39.6 & 19886.7 & 0 & 0 & 0 & 0 & 0 & 3 & 0 & 0 & 48/50 \\
6 & 0.98 & 1.00 & 29.0 & 8493.8 & 0 & 0 & 0 & 0 & 0 & 0 & 2 & 0 & 49/50 \\
7 & 0.98 & 1.00 & 19.2 & 16032.2 & 0 & 0 & 0 & 0 & 0 & 1 & 0 & 0 & 49/50 \\
8 & 0.90 & 0.99 & 53.4 & 22461.3 & 0 & 0 & 0 & 0 & 3 & 2 & 0 & 0 & 45/50 \\ \midrule
OVERALL & 0.97 & 1.00 & 30.2 & 13390.9 & 0 & 2 & 0 & 0 & 3 & 6 & 2 & 0 & 389/400 \\
\bottomrule
\end{tabular}

\vspace{1mm}
\begin{minipage}{0.98\textwidth}
\footnotesize
\emph{Note:} Error classification columns count discrepancy instances rather than failed runs. A single failed output may contain multiple errors, including more than one error of the same type.
\end{minipage}
}

\end{table}

\begin{table}[h]
\centering
{\footnotesize
\caption{Stage 2 Evaluation Results by case$^*$}
\label{tab:stage2_results}
\begin{tabular}{lcccc}
\toprule
 Case ID / Model Name & Pass@1 & Passes & Latency (s) & Tokens \\
\midrule
1 and 3 & 1.0 & 20/20 & 4.4 (1.5) & 2733 (110) \\
2 & 1.0 & 20/20 & 8.4 (3.2) & 3479 (93) \\
4 & 1.0 & 20/20 & 10.6 (4.2) & 3712 (95) \\
5 & 1.0 & 20/20 & 2.7 (0.8) & 2433 (98) \\
6 & 1.0 & 20/20 & 4.9 (1.5) & 2870 (115) \\
7 & 1.0 & 20/20 & 6.8 (2.4) & 3031 (102) \\
8 & 1.0 & 20/20 & 16.9 (6.0) & 4404 (148) \\
\midrule
gemma-4-26B-A4B-it & 1.0 & 80/80 & 4.8 (2.8) & 3271 (624) \\
Mistral-Small-3.2-24B-Instruct-2506 & 1.0 & 80/80 & 9.9 (5.9) & 3067 (606) \\
\bottomrule
\end{tabular}
\begin{tablenotes}
\centering
\item $^*$Latency and Tokens are mean (Std) figures.
\end{tablenotes}
}
\end{table}

\begin{table}[htbp]
  \centering
  {\footnotesize
  \caption{Performance of end to end workflow: overall and across evaluation cases$^*$}
  \label{tab:end2end_results}
  \begin{tabular}{l r r r r r}
    \toprule
    \textbf{Case ID} & \textbf{Passes} & \textbf{Pass@1} & \textbf{Pass@2} & \textbf{Latency (s)} & \textbf{Tokens} \\
    \midrule
    1 & 50/50 & 1.00 & 1.00 & 21 (7) & 9258 (1403) \\
    2 & 50/50 & 1.00 & 1.00 & 37 (8) & 12914 (1492) \\
    3 & 50/50 & 1.00 & 1.00 & 20 (5) & 9053 (1272) \\
    4 & 47/50 & 0.94 & 1.00 & 41 (9) & 23025 (1933) \\
    5 & 47/50 & 0.94 & 1.00 & 42 (11) & 22789 (2091) \\
    6 & 49/50 & 0.98 & 1.00 & 32 (8) & 11625 (1538) \\
    7 & 50/50 & 1.00 & 1.00 & 23 (6) & 19354 (1165) \\
    8 & 45/50 & 0.90 & 0.99 & 65 (12) & 27514 (2280) \\
    \midrule
    OVERALL & 388/400 & 0.97 & 1.00 & 35 (16) & 16942 (6873) \\
    \bottomrule
  \end{tabular}
  \begin{tablenotes}
  \centering
\item  $^*$Latency and Tokens are mean (Std) figures rounded to nearest integer. \item Pass@k rounded to 2dp.
\end{tablenotes}
  }
\end{table}
\section{Discussion}

This study tested whether free and open-source software, combined with open-weight LLMs, could be used to convert diagrams of open queuing networks with reneging into verifiable computer simulation models. We proposed and evaluated a three-stage workflow: first, a multimodal LLM translates an image of a queuing-network diagram into a semi-structured textual description; second, an LLM translates this description into schema-validated JSON; third, our own rule-based software \textit{json2ciw} automatically converts the generated JSON model data into a \textit{Ciw} simulation model. The central design choice was to avoid direct generation of simulation code and instead to create an intermediate, inspectable model representation that can be automatically validated and reviewed by a human modeller. We emphasise that our automatic schema validation ensures that the JSON is structurally valid and conforms to allowable distributions, parameters, and routing constraints.  However, a schema-valid specification is not necessarily conceptually correct. A human-in-the-loop approach is still essential to ensure the model has been translated correctly.

Across the eight evaluation cases, our workflow showed high reliability. End-to-end, the workflow generated the expected \textit{Ciw} model in 388 of 400 runs, giving an overall $Pass@1$ of 0.97 and Pass@2 of 1.00. These results suggest that, for the class of models evaluated here, an open-weight LLM workflow can generate usable simulation-model specifications with a high probability of success, while preserving intermediate artefacts that support verification.

We acknowledge that use of LLMs in the simulation modelling process introduces additional energy usage above and beyond traditional software tools and there is growing consensus that carbon footprint should be considered in LLM research \citep{JEANQUARTIER_EnvImpact_LLM}. On average, our workflow took 35 seconds to run on a single NVIDIA H100 GPU, with 80 GB of allocated memory. We did not measure energy usage directly, but a conservative estimate can be given using the \cite{Lannelongue_Carbon_calculator} method that adjusts for GPU type, cloud versus on-site server location, and the carbon intensity of the national electricity grid. Assuming full GPU VRAM utilisation during a run, the workflow was estimated to consume an average of 5.73 Wh. In the United Kingdom, this corresponds to a carbon footprint of 939 mgCO$_2$e (milligrams of CO$_2$ equivalent), equivalent to $1.02 \times 10^{-3}$ tree-months, where one tree-month represents the volume of CO$_2$ absorbed by a single mature tree over one month\footnote{Calculated using \url{green-algorithms.org} v3.1; \cite{Lannelongue_Carbon_calculator}}.  Case 8, our longest running test, took 65s on average with an estimated carbon footprint of 1.8 gCO$_2$e.

\subsection{Failure modes}

The main source of error we observed was in the image-to-description stage. Failures were concentrated in the more complex cases, particularly those involving the larger networks and routing loops. The most common discrepancies involved routing and parameter interpretation, rather than hallucinated activities or complete generation failures. 

Routing errors occurred in hand-drawn networks with nodes that looped back to prior nodes (cases 5, 7 and 8). In simple terms the LLM misinterpreted where the edges were pointing. It is unclear whether this was due to the proximity of lines or nodes to each other. 

Case 8, the largest network, was the most challenging for our workflow. In addition to routing errors the lower reliability is also explained by case 8 having 2 external arrival sources. In a small number of cases the LLM generated a node arrival rate equal to the outflow of a prior node with the external arrival rate. In contrast, Case 2, a computer-drawn Jackson network with three external arrival sources and multiple routing loops, achieved perfect reliability across all 50 runs. One possible explanation is an interaction between the hand-drawn presentation and the higher node count.

\subsection{Implications for the workflow}

The pattern of success and failures in our results demonstrates that Stages 2 and 3 of the workflow can be robust once a faithful description has been obtained, but that visual interpretation remains an area where the most caution is needed. In practical terms, this supports a workflow in which users inspect the generated description, process diagram, parameter tables, or JSON representation before accepting the model as a basis for simulation. Another approach, for more complex diagrams, is to generate two simulation models, and intermediate artefacts, in parallel, as in our test cases we observed that at least one of generated models was correct with a high probability. Our workflow could be adapted to compare the two models. If two independent runs produce different simulation artefacts then the workflow could prompt for human review of specific details and quickly select the correct translation.  

A second implication is that Stage 3 should be understood as an adapter layer rather than as a commitment to a single simulation package. In this study we used \textit{Ciw} as the target simulation package and implemented the \textit{json2ciw} adapter to convert JSON model specifications into executable simulation models. This choice reflected the aims of the study: \textit{Ciw} is free and open-source, has a concise Python interface, and is well suited to representing open queuing networks. It also supports the wider reproducibility aims of the workflow, because the model-generation process, intermediate representation, conversion rules, and simulation implementation can all be inspected and re-run.  The separation of the adapter creates a route for extending the workflow beyond \textit{Ciw}. The JSON representation is intended to act as a package-agnostic intermediate model specification, with the simulation package treated as the target of a deterministic adapter rather than as the centre of the workflow. Stage 3 could therefore be implemented for other simulation tools by developing alternative adapters. These could include adapters for other free and open-source tools, such as \textit{SimPy} or \textit{simmer}, or import routines for commercial simulation packages such as \textit{Simio}, \textit{Simul8}, or \textit{AnyLogic}. This has practical implications for adoption: organisations committed to reproducible open-source workflows could use FOSS adapters, while organisations already invested in commercial simulation software could potentially use the same upstream \textit{Sketch2DES} workflow with a different Stage 3 adapter.  This flexibility does come at a maintenance cost. The reliability of the  adapter is achieved because it is a set of deterministic rules implemented in software. As the JSON schema representing models is expanded, or the simulation package evolves, so must the adapter.

The architecture we have designed also positions our work differently from approaches that integrate LLM functionality directly within a specific simulation environment, such as SimGPT with \textit{Simio} \citep{DEHGHANI_SIMGPT}. Our emphasis is on generating an intermediate representation of a DES model that, in theory, can be mapped to many simulation packages. We also aim to allow users more flexibility in the diagrams of queuing networks they use, including hand-drawn diagrams, and as such do not require diagrams that are specific to a simulation package such as Simulink \citep{ren2025simugen}.

\subsection{Infrastructure and accessibility}

A motivation for using open-weight LLMs and free and open-source software is that the workflow should not depend on commercial API access or externally hosted AI services. This is important for reproducibility, because commercial model interfaces, pricing, licensing terms, and underlying model behaviour may change over time. It is also important for privacy-sensitive organisations, including those working with health-service processes, where data governance requirements may restrict the use of externally hosted AI systems. Local deployment therefore forms part of the wider methodological rationale for the workflow, rather than being only an implementation detail.

The formal evaluation was conducted on institutional GPU infrastructure because it provided the compute environment required for repeated runs, latency measurement, and comparison across evaluation cases. However, a motivation for using open-weight models and free and open-source software is that our workflow should not be restricted to HPC infrastructure. Informally, we confirmed that the workflow can be executed end-to-end on a desktop workstation equipped with an NVIDIA RTX 5090 GPU (hardware typically used for playing video games), using quantised versions of the Qwen and Gemma LLMs and the Ollama inference engine. Quantisation reduces the memory and computational requirements of an LLM by representing model weights at lower numerical precision, in our case 4-bit values rather than full-precision floating point. This suggests that private, local \textit{Sketch2Sim} workflows may be technically feasible beyond HPC environments.  We included the code to run the LLM workflow on a desktop PC in our archived code.

Our desktop deployment test is not intended as a second reliability evaluation. The reliability results reported in this study are therefore based on the formal experimental setup described above. Quantisation, GPU memory, and inference engine configuration, may all affect latency and, potentially, model outputs. A systematic comparison of full-precision and quantised deployments across HPC and desktop hardware is an important area for future work, particularly if the method is to be adopted by analysts or organisations without access to high-end GPU infrastructure.

\subsection{Comparison with direct code generation approaches}

A key motivation for this work was to address limitations observed in prior approaches based on direct LLM generation of simulation code \citep{Monks_Unlocking2025,jackson_2024,schmitt2026llm}. Taken with our study, we can now arrange these contributions as a continuum of model-generation approaches. At one extreme we have direct use of LLM workflows to translate natural language descriptions directly into executable code \citep{Monks_Unlocking2025,jackson_2024}. This approach is the most flexible, but also introduces the highest risk of mistakes and the highest user knowledge and skill requirements. \cite{schmitt2026llm}'s approach is a mid-point on the continuum, and aims to improve the reliability of GenAI by blending pre-written blue-print model code with LLM code writing capabilities. This anchors the generated models in a specific domain to increase reliability, but allows adaptation and extension via code generation, and hence still suffers from LLM hallucination and mistakes although less than pure code generation.  Our new study sits at the other end of the continuum. Our approach constrains the LLM to narrow translation tasks and does not generate any simulation model code. Our approach is therefore similar to the use of LLMs as smart translators proposed by \cite{Giabbanelli_translators} and takes \cite{schmitt2026llm}'s determinism further to substantially increase reliability. The approach can be applied to any queuing domain, and requires less direct programming knowledge, but is limited by what has been defined in the Stage 2 simulation schema, the rules in the Stage 3 simulation package adapter, and the capability of the Stage 1 vision LLM. 

\subsection{Strengths}

Our approach was motivated by the longer-term aim of lowering the programming barrier for simulation users who have high system and data knowledge, but potentially limited model-building skills. Such users might include analysts who work in health services. By embedding the method in the emerging \textit{Sketch2Sim} paradigm, the workflow provides a technical foundation for model development that begins with visual representations of process logic and data assumptions, rather than with simulation code. In a future applied setting, this could support problem-structuring meetings in which a process is sketched, photographed, uploaded, and converted into a draft simulation model for human review. However, the present study does not evaluate this end-user workflow directly. Our results should therefore be interpreted as evidence for the feasibility and reliability of the underlying technical architecture, rather than as evidence that the approach has lowered barriers for novice modellers in practice.

For the eight case study models included in our evaluation, we observed high reliability and speed on private infrastructure using open-weight models. Although only a small number of cases were used, a strength of our evaluation is that the sample is larger than those in several existing studies of generated simulation models, which typically use no more than one run of one or two case studies \citep{Monks_Unlocking2025,jackson_2024,schmitt2026llm}. Another strength of our evaluation methodology is that it measures both accuracy and reliability through multiple runs of each case and the Pass@k metric. We therefore believe that our results provide evidence that this architecture is a promising foundation for more transparent GenAI-assisted model building, with lower risks than direct code generation approaches.

A further strength of our study is the use of only open-weight LLMs. This methodology offers considerable advantages for transparency and reproducibility compared with existing studies, including our own initial work, that used proprietary LLMs. Not only can our approach be recreated in academic and industry settings, but we can also re-run the workflow at a future date with the same models, updated prompts, or different open-weight LLMs as the technology progresses.

\subsection{Limitations}

The objective of this study was to establish the feasibility of a staged, verifiable workflow architecture, rather than to demonstrate fully-automated simulation model generation. The queuing-network examples therefore serve as a proof-of-concept testbed for evaluating the reliability of the approach. Although we used a variety of diagrams with differing levels of complexity and notation, our evaluation remains small and may not be fully representative of process flow diagrams drawn in practice. Real stakeholder-generated diagrams may be incomplete, ambiguous, inconsistent, or use informal notation that differs from the diagrams included in this evaluation. The results should therefore be interpreted as evidence that the workflow is feasible for a defined class of open queuing-network diagrams, rather than as evidence that it can automatically formalise arbitrary stakeholder sketches into simulation models.

An opportunity for future research would be to consider image pre-processing before passing to the Stage 1 LLM. In our experiments, the only pre-processing we conducted was to rotate the image, if required, so that it was horizontal. This meant that we passed varying sizes of image to the LLM. Images could be standardised in size or enhanced prior to prompting an LLM to describe it.

The current schema and adapter also support only a restricted class of open queuing-network models, including arrivals, service processes, routing, resources and reneging. They do not yet represent features such as resource schedules, priorities, batching, shared resources or state-dependent logic. The findings therefore cannot be assumed to extend to more complex DES structures without corresponding extensions to the schema and software adapters.

Finally, because the evaluation used a fixed and documented experimental design, using specific LLM models and GPU hardware, the results should be interpreted as conditional on that configuration. This supports reproducibility, but does not by itself establish robustness across all possible model families, inference engines, hardware environments, quantised deployments, or diagram sources. We note that we did observe some robustness in Stage 2: both Gemma 4 and Mistral Small 3.2 generated the expected structured model representations in all tested cases, with Gemma 4 providing lower latency. However, this comparison was limited to the text-to-JSON stage, and the main source of end-to-end error in our evaluation was the multimodal image-to-description stage. Further work is therefore needed to evaluate whether similar reliability is observed across alternative multimodal models, deployment environments, and diagrams produced by users outside the study team.

\section{Conclusion}

This study explored the feasibility of using open-weight LLMs and free and open-source software to convert diagrams of open queuing networks into verifiable simulation models. Beyond demonstrating feasibility, the study introduces a staged workflow in which intermediate, software-independent model specifications can be inspected and verified before deterministic model construction. This contrasts with direct code generation approaches by making each transformation step transparent and independently verifiable. 

Our findings demonstrate the feasibility of this approach for the class of models evaluated here, with the proposed three-stage workflow generating the expected \textit{Ciw} model in 388 of 400 end-to-end executions. The results show the value of using LLMs for constrained translation tasks, from image to description and from description to structured JSON, while reserving final model construction for deterministic software. This design creates intermediate artefacts (data tables and diagrams) that support human verification and reduces some of the risks associated with direct simulation code generation. However, the study also highlights important limitations. Errors were concentrated in the visual interpretation stage, particularly for larger hand-drawn networks with routing loops and multiple arrival sources.  

Further research is needed to evaluate the workflow with real stakeholder-generated diagrams, extend the JSON schema and adapters to more complex classes of simulation models, and assess deployment across alternative hardware, quantised models, and target simulation packages. More broadly, the staged architecture provides a foundation for reproducible, transparent and verifiable AI-assisted simulation model generation across a wider range of simulation domains and software platforms. 

\section*{Code availability}

\noindent The pilot evaluation is available on GitHub \href{https://github.com/sim-agent/sketch-to-des-pilot}{https://github.com/sim-agent/sketch-to-des-pilot} and archived in Zenodo \citep{monks_sketch2des_pilot}. Our ciw adapter package \textit{json2ciw} can be installed from Github; we used version 0.10.0 for the evaluation \href{https://github.com/sim-agent/json2ciw}{https://github.com/sim-agent/json2ciw} and archived on Zenodo \citep{monks_json2ciw}.

\section*{Disclosure statement}

The authors confirm there is no conflict of interest.

\section*{Funding}

This work was supported by the Medical Research Council [MR/Z503915/1].

\section*{CRediT}
Conceptualization: TM; Data curation: TM; Formal Analysis: TM; Funding acquisition: TM, AH, NM; Investigation: TM, AH; Methodology: TM; Project administration: TM, AH; Resources: TM, AHe; Software: TM, AH, AHe; Supervision: TM; Validation: TM, AH, AHe; Visualization: TM, AH; Writing – original draft: TM, AH, AHe, NM; Writing – review \& editing: TM, AH, AHe, NM

\bibliographystyle{apalike} 

\bibliography{refs}

\pagebreak
\begin{appendices}
\counterwithin{figure}{section}

\section{LLM Workflow Prompts}
\label{app:prompts}

\subsection{Stage 1 Prompt}
\label{app:stage1_prompt}
\begin{verbatim}
You are an expert in analyzing process flow diagrams and discrete event simulation models.
Your task is to analyze the attached process flow diagram image and extract a detailed,
structured description.

ANALYSIS REQUIREMENTS:

1. Identify all activities/processes in the diagram.

2. Extract service time distributions with exact parameters
   (e.g., normal(mean=15, sd=3), lognormal(mean=2.5, sd=0.8), exponential(mean=0.25)).
  
3. For EACH activity, determine whether external entities arrive        
   into it OR into its waiting queue. 
    - If an external source feeds an activity or its queue, 
      extract that as the activity's Arrival distribution. 
    - Only output 'none - internal routing only' if the activity exclusively receives 
    entities from other upstream activities.

4. For EACH activity, determine whether the diagram explicitly shows a reneging,
   abandonment, patience, timeout, or maximum waiting-time distribution associated
   with that activity or its queue.
   - If yes, extract it exactly as a reneging distribution.
   - If not shown, do NOT infer one and do NOT include one.

5. For exponential distributions, determine carefully whether the parameter
   is a RATE or a MEAN (inter-arrival times imply MEAN, arrival rates imply RATE → convert).
   e.g., "rate=3/hour → mean=1/3=0.333 hours"

6. Evaluate ALL mathematical expressions to their decimal value before inclusion.
   Examples: 1/3 → 0.3333, 10*2 → 20.

7. Note resources associated with each activity (name and capacity).

8. Ensure all distributions use consistent time units throughout.

9. Exits/sinks are NOT activities. Always refer to them as EXIT. 
   Do not include in the list of Activities.

10. Arrival/inter-arrival time distribution nodes are NOT activities.

11. Reneging/abandonment/patience distribution nodes are NOT activities.

12. Decision points without a time are NOT activities.

13. Extract a routing matrix between ALL activities and EXIT that follows the logic 
    in the diagram exactly.
    - Do not use your own knowledge of processes to infer what the user should have drawn.
    - Your final description of this must only include Activities and Exits.
    I.e. Do not include arrival/service/reneging distributions, resources or nodes 
    that act as splitters.
    - Ensure all edge probabilities sum to 1.0 for each node. Use exact activity names.
    - Perform a final double check that only Activities and EXITs are 
    included in the routing matrix.

14. For each link between activities pay careful visual attention to the arrowheads 
    of intersecting lines and feedback loops to ensure you do not reverse directions. 

15. If any information is ambiguous, illegible, or missing, state this explicitly.

OUTPUT FORMAT:

Provide a structured description organised exactly as follows.

**Process Overview:**
[Brief description of what the process represents and its overall structure]

**Activities:**
- Activity 1: [name]
  - Service distribution: [type(param=value, ...)]
  - Arrival distribution: [type(param=value, ...)] OR "none - internal routing only"
  - Reneging distribution: [type(param=value, ...)] OR "none"
  - Resource: [resource name] (capacity=[N])

[continue for ALL activities]

**Activity Routing:**

- Routing 1 (after [activity name]):
  → Branch A (probability: 0.X) → [destination activity name or EXIT]
  → Branch B (probability: 0.X) → [destination activity name or EXIT]

- Routing 2 (after [activity name]):
  → Branch A (probability: 1.0) → [EXIT]

[continue for ALL activities]

**Flow Logic:**
[Step-by-step narrative of how entities move through the system from arrival to exit.]
\end{verbatim}

\subsection{Stage 2 prompt template}
\label{app:stage2_prompt}

\begin{verbatim}

{{ json_schema }}

You are an expert at converting process descriptions into structured JSON 
for discrete event simulation models.

### Process Overview

{{ process_description }}

CRITICAL RULES:

1. **EVALUATE ALL EXPRESSIONS**: 
   - ANY mathematical expressions (e.g., 60/200, 1/2, 10*2) MUST be evaluated 
   to their decimal value before inclusion. Do the calculation step by step.
   - Examples: 1/2 → 0.5, 10*2 → 20

2. **STANDARD PARAMETER NAMING**:
   - Use standard parameter names: 'min', 'max', 'mean', 'sd', 'rate', 'mode'
   - NEVER use synonyms like 'minimum', 'maximum', 'average', 'standard_deviation'
   - Preserve exact parameter names as they appear in the description
   - NEVER use numbers as parameter names (e.g., "5", "7", "10" are invalid keys).

3. **NODE CONSISTENCY AND EXIT NODES**:
   - The "from" field in transitions must be an activity name from the "activities" array.
   - The "to" field in transitions must be either an activity name from the 
   "activities" array or "Exit".
   - DO NOT reference or create other sink or end node names such as "Start", "End", 
   "call ended", "finished", "completed", etc.
   - When a process leaves the system after an activity, ALWAYS use "Exit" as the "to" value.
   - Do NOT add "Exit" to the "activities" array unless it is explicitly defined as 
   an activity in the description.

4. **INTER-ARRIVAL TIME HANDLING**:
   - THERE MUST BE AT LEAST ONE ACTIVITY with an inter-arrival process in the model.
   - Prefer assigning the inter-arrival distribution to the first activity in the flow 
   (e.g., "call triage" for arrival of calls).
   - If inter-arrival time is NOT SPECIFIED anywhere in the description, 
   you MUST still create an inter-arrival distribution for exactly one activity, using:
     {
       "type": "exponential",
       "parameters": {
         "rate": 1.0
       }
     }
   - Never leave all "arrival_distribution" fields as null.
    
5. **PROBABILITY VALIDATION**:
   - Ensure ALL probabilities are decimal numbers between 0.0 and 1.0
   - For each source node, all outgoing transition probabilities must sum to exactly 1.0

6. **STRICT JSON COMPLIANCE**:
   - Output ONLY valid JSON with no additional text
   - Use proper JSON syntax (commas, brackets, quotes)
   - No trailing commas, comments, or explanations

7. **DISTRIBUTION NOTATION**:
   - Use parameter names exactly as supplied in the description.
   - If the description omits parameter names but gives values in a standard order,
     you MUST infer the correct names rather than using raw numbers as keys.
   - Examples of standard mappings:
     - triangular(a, b, c) → "min": a, "mode": b, "max": c
     - uniform(a, b) → "min": a, "max": b
     - lognormal(m, s) where the description says "mean" and "sd"
       → "mean": m, "sd": s
   - NEVER use numeric strings as keys in the "parameters" object
     (e.g., "5", "7", "10" are invalid parameter names).

OUTPUT INSTRUCTIONS:
- Begin with the opening curly brace
- End with the closing curly brace
- Include NO text before or after the JSON
\end{verbatim}

\section{LLM-as-a-judge}
\label{app:llm-judge}

\subsection{JSON schema used by Judges}

{\footnotesize
\begin{verbatim}
    

{
  "description": "Evaluates a generated process description against a ground-truth rubric.",
  "properties": {
    "reasoning": {
      "description": "Detailed step-by-step reasoning evaluating Activities, Distributions, Parameters, Resources, 
      and Routing. Identify specific errors before tallying them below.",
      "title": "Reasoning",
      "type": "string"
    },
    "misclassified_nodes": {
      "description": "COUNT of nodes incorrectly listed as an Activity. CRITICAL: Use this category 
      if the concept/text actually exists SOMEWHERE in the Ground Truth (like a Decision, Arrival, or Exit), 
      but was just given the wrong label. Do not count parameter/resource errors for these.",
      "title": "Misclassified Nodes",
      "type": "integer"
    },
    "hallucinated_nodes": {
      "description": "COUNT of completely invented activities. CRITICAL: Use this category ONLY 
      if the concept/text does NOT exist ANYWHERE in the Ground Truth. If it exists as a Decision/Arrival, 
      it is Misclassified, NOT Hallucinated. Do not count parameter/resource errors for these.",
      "title": "Hallucinated Nodes",
      "type": "integer"
    },
    "missed_nodes": {
      "description": "COUNT of activities in the Ground Truth that are completely missing 
      from the generated Activities list.",
      "title": "Missed Nodes",
      "type": "integer"
    },
    "missed_distributions": {
      "description": "COUNT of distributions in the Ground Truth that are completely missing 
      from the generated description including arrivals, service and renege distributions.",
      "title": "Missed Distributions",
      "type": "integer"
    },
    "parameter_errors": {
      "description": "COUNT of incorrect distribution parameters (e.g. wrong mean, wrong min/max, 
      wrong distribution name).",
      "title": "Parameter Errors",
      "type": "integer"
    },
    "routing_errors": {
      "description": "COUNT of incorrect routing probabilities, missing branches, or probabilities 
      that fail to sum to 1.0.",
      "title": "Routing Errors",
      "type": "integer"
    },
    "resource_capacity_errors": {
      "description": "COUNT of incorrect integer capacities (N) assigned to resources.",
      "title": "Resource Capacity Errors",
      "type": "integer"
    },
    "generation_error": {
      "description": "1 if the LLM failed to generate a valid process description e.g. ran out of tokens 
      during thinking;If one then ALL other categories of error should be set to 0; 0 otherwise.",
      "title": "Generation Error",
      "type": "integer"
    },
    "score": {
      "description": "1 if perfectly matches Ground Truth (0 errors). 0 if ANY errors exist.",
      "title": "Score",
      "type": "integer"
    }
  },
  "required": [
    "reasoning",
    "misclassified_nodes",
    "hallucinated_nodes",
    "missed_nodes",
    "missed_distributions",
    "parameter_errors",
    "routing_errors",
    "resource_capacity_errors",
    "generation_error",
    "score"
  ],
  "title": "DiagramJudgement",
  "type": "object"
}

\end{verbatim}
}

\subsection{Judge prompt}

\begin{verbatim}
You are an expert evaluator grading an AI's ability to extract discrete event 
simulation data from a process flow diagram.

You will compare a 'Generated Description' against a 'Ground Truth Description'. 
The generated text follows a structured format containing **Activities** 
and **Activity Routing**

Your job is to find any discrepancies and categorize them strictly into eight error types 
using the provided JSON schema.

CRITICAL EVALUATION RULES:
- **Routing Logic**: 
Extract and compare routing strictly as unordered {Destination: Probability} pairs 
(e.g., {Service 1: 0.4, Service 2: 0.4}). 
Ignore arbitrary branch labels and list order. 
Mathematical equivalence means there is NO error. Check if probabilities fail to sum to 1.0.

SCORING RULE:
- Score 1 ONLY if the generated extraction perfectly matches the Ground Truth logic 
(ignoring minor formatting differences).
- Score 0 if you find ANY hallucination, omission, misclassification, resource error, 
or parameter/routing error.

Always write your step-by-step `reasoning` checking the Activities, Distributions, 
Parameters, Resources, and Routing before you output the final score and error lists.

## Ground Truth Description
{{ ground_truth }}

## Generated Description (to evaluate)
{{ generated }}

\end{verbatim}

\section{Images used in evaluation}

\begin{figure}
    \centering
    \includegraphics[width=0.75\linewidth]{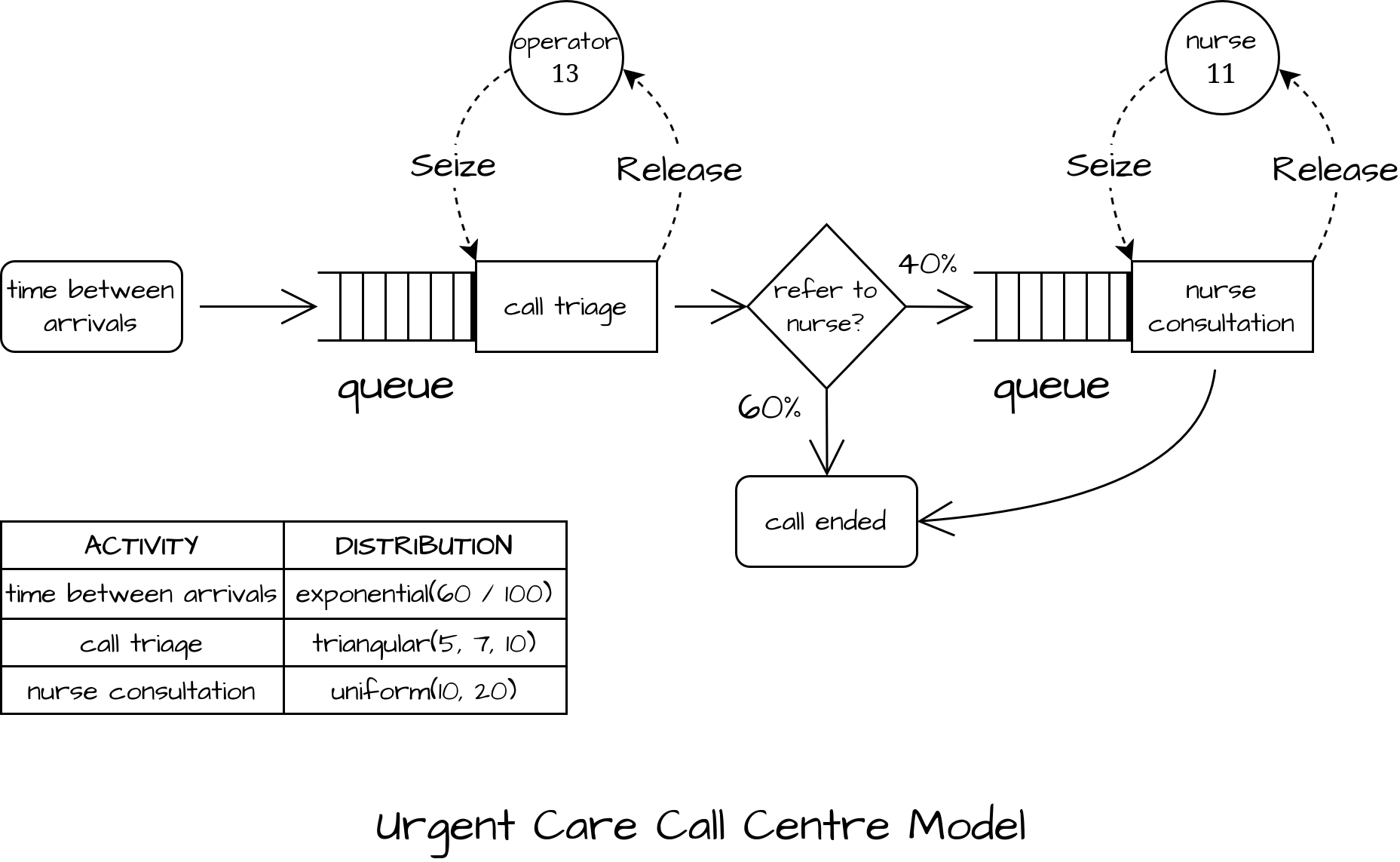}
    \caption{Case 1: Call Centre}
    \label{fig:case1_img}
\end{figure}

\begin{figure}
    \centering
    \includegraphics[width=0.75\linewidth]{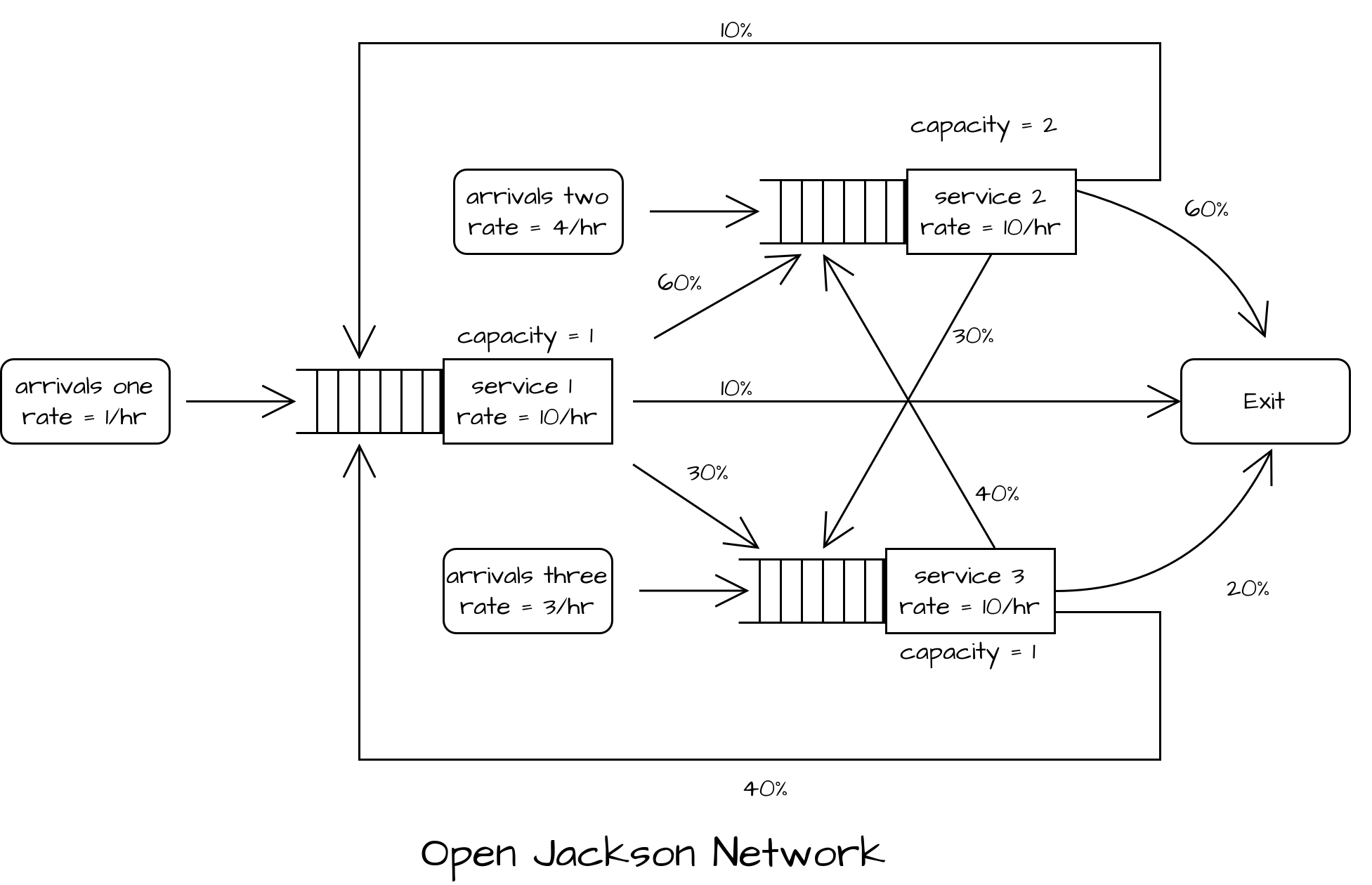}
    \caption{Case 2: Jackson Network 1}
    \label{fig:case2_img}
\end{figure}

\begin{figure}
    \centering
    \includegraphics[width=0.75\linewidth]{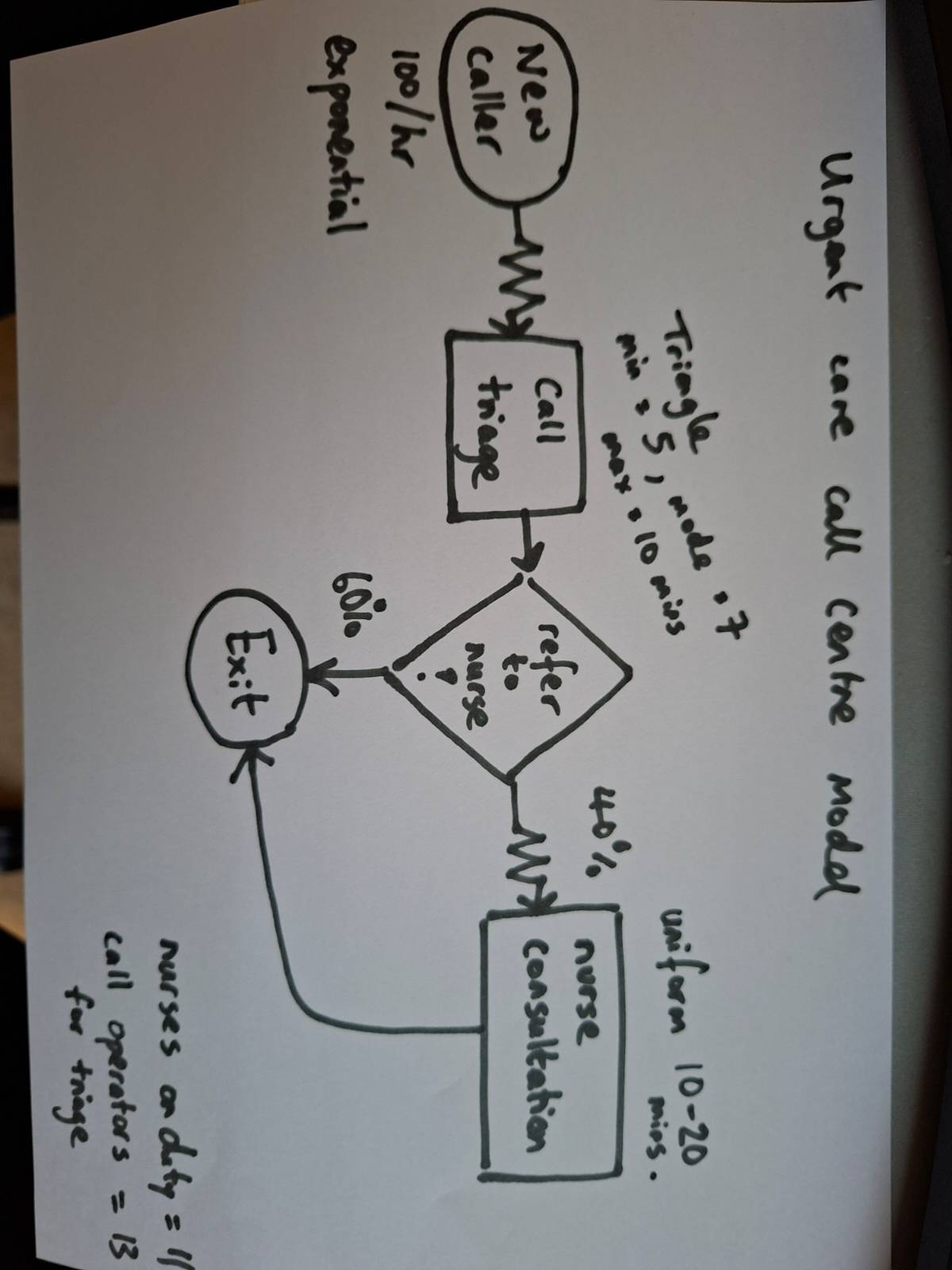}
    \caption{Case 3: Hand drawn call centre}
    \label{fig:case3_img}
\end{figure}

\begin{figure}
    \centering
    \includegraphics[width=0.75\linewidth]{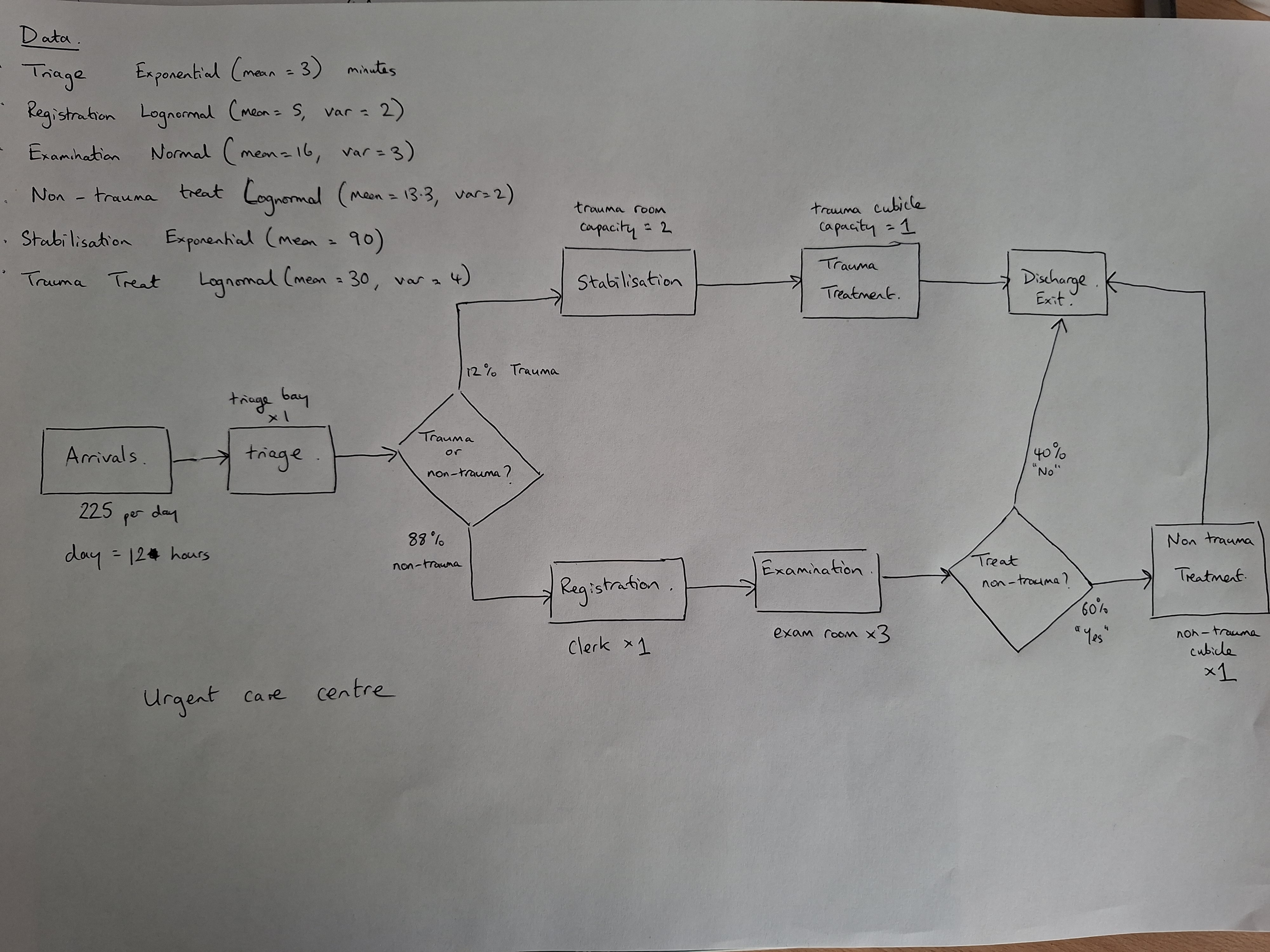}
    \caption{Case 4: Treatment centre}
    \label{fig:case4_img}
\end{figure}

\begin{figure}
    \centering
    \includegraphics[width=0.75\linewidth]{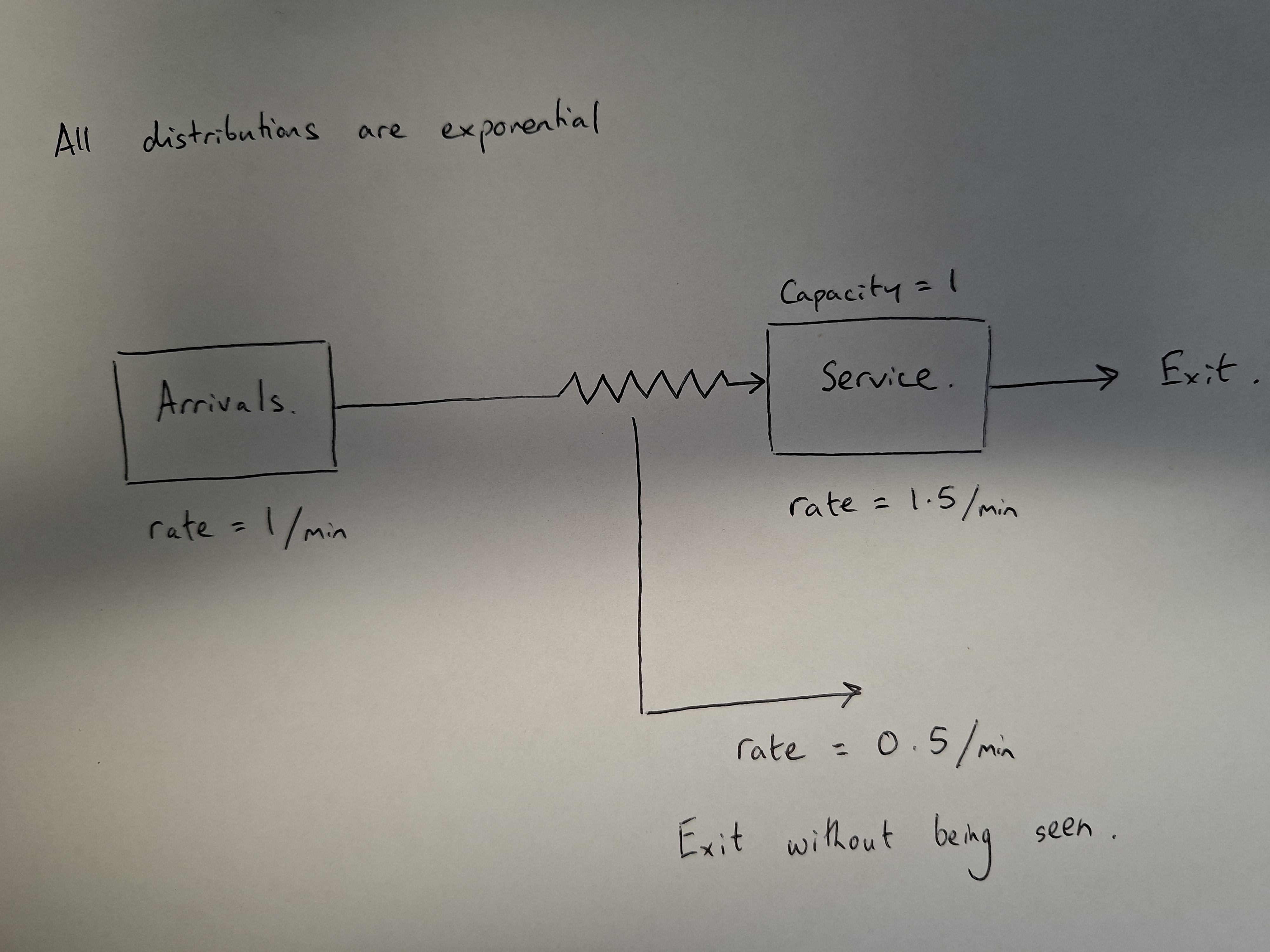}
    \caption{Case 5: MM1 with renege}
    \label{fig:case5_img}
\end{figure}

\begin{figure}
    \centering
    \includegraphics[width=0.75\linewidth]{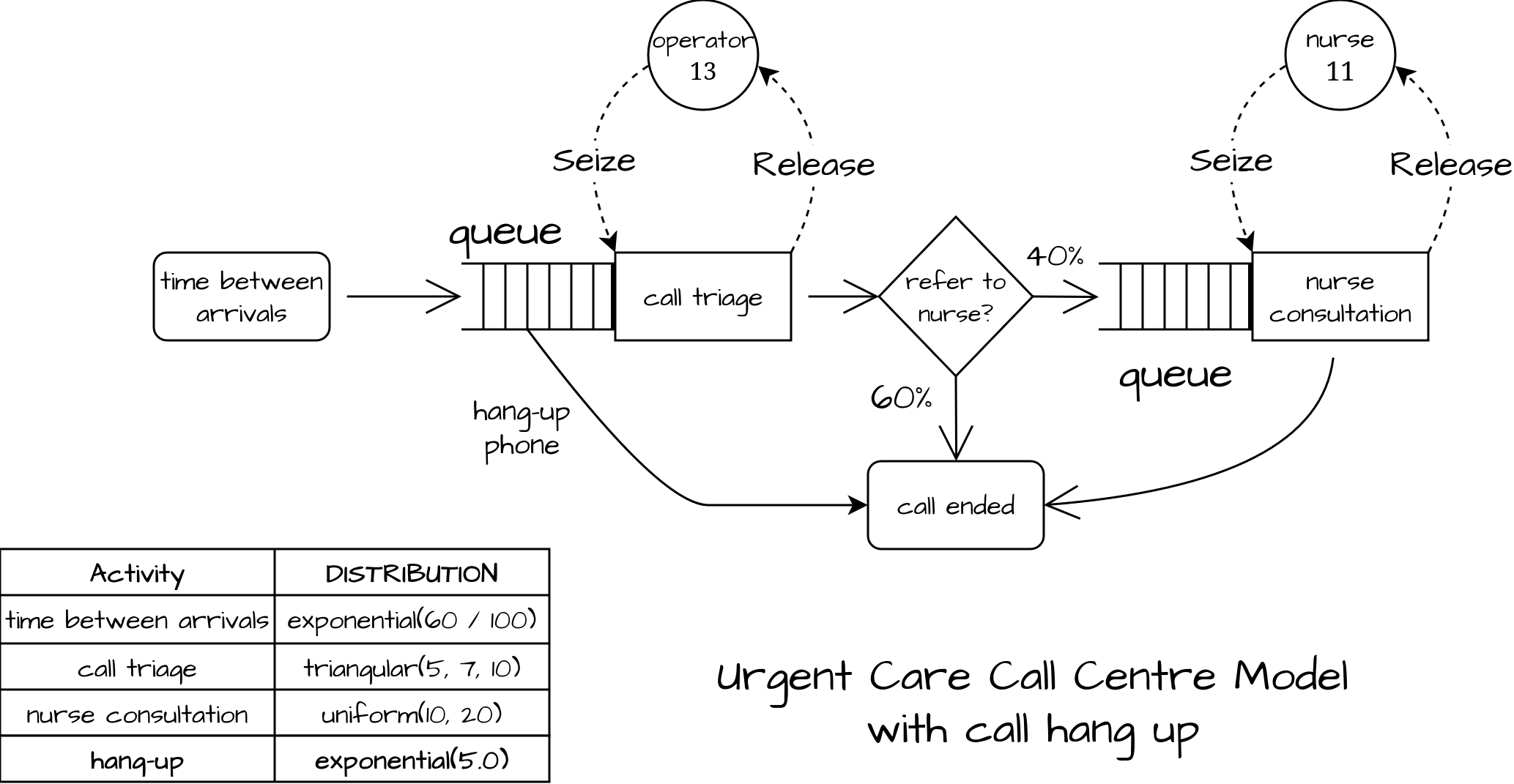}
    \caption{Case 6: Call centre renege}
    \label{fig:case6_img}
\end{figure}

\begin{figure}
    \centering
    \includegraphics[width=0.75\linewidth]{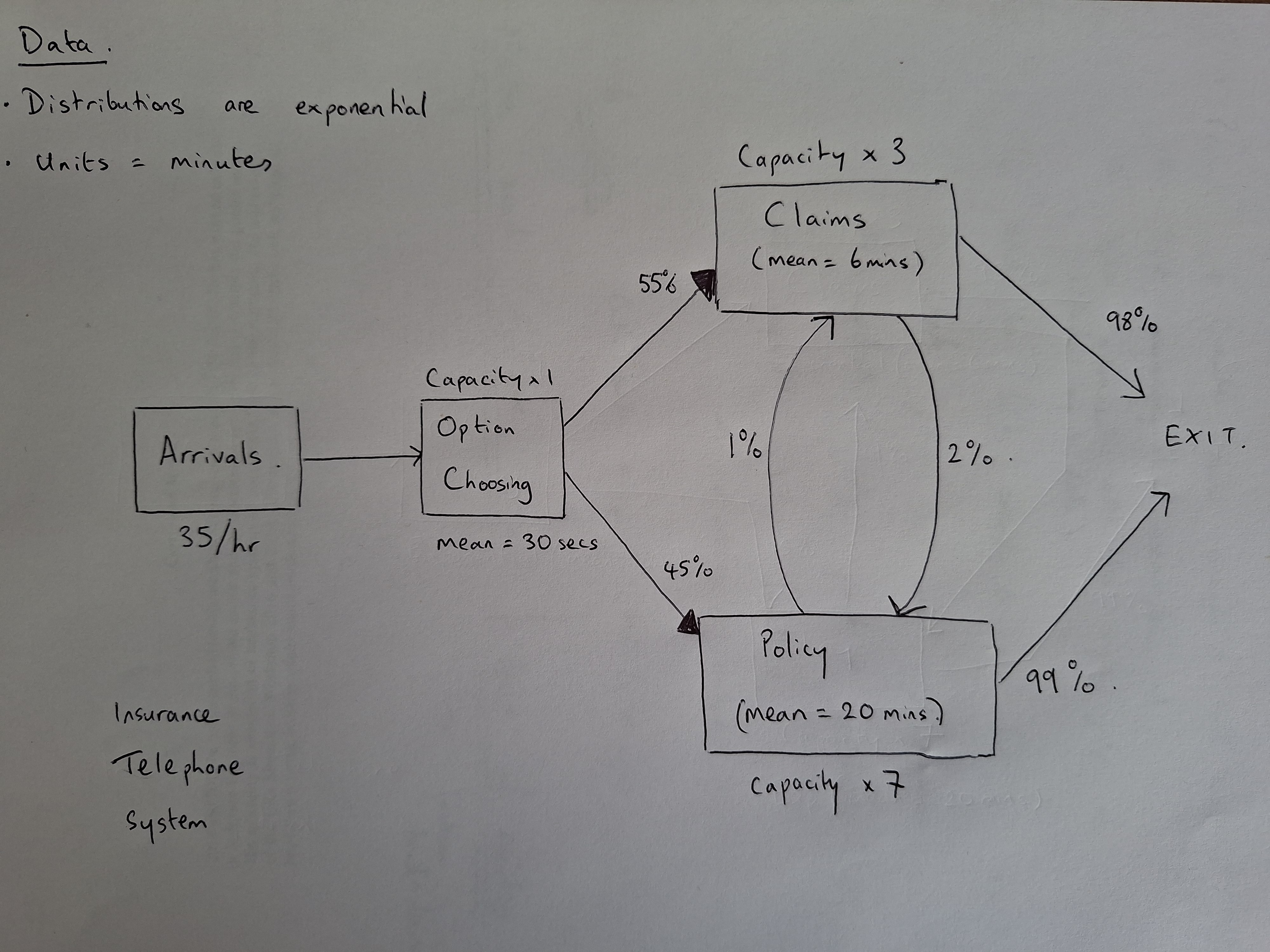}
    \caption{Case 7: Jackson Network 2}
    \label{fig:case7_img}
\end{figure}

\begin{figure}
    \centering
    \includegraphics[width=0.75\linewidth]{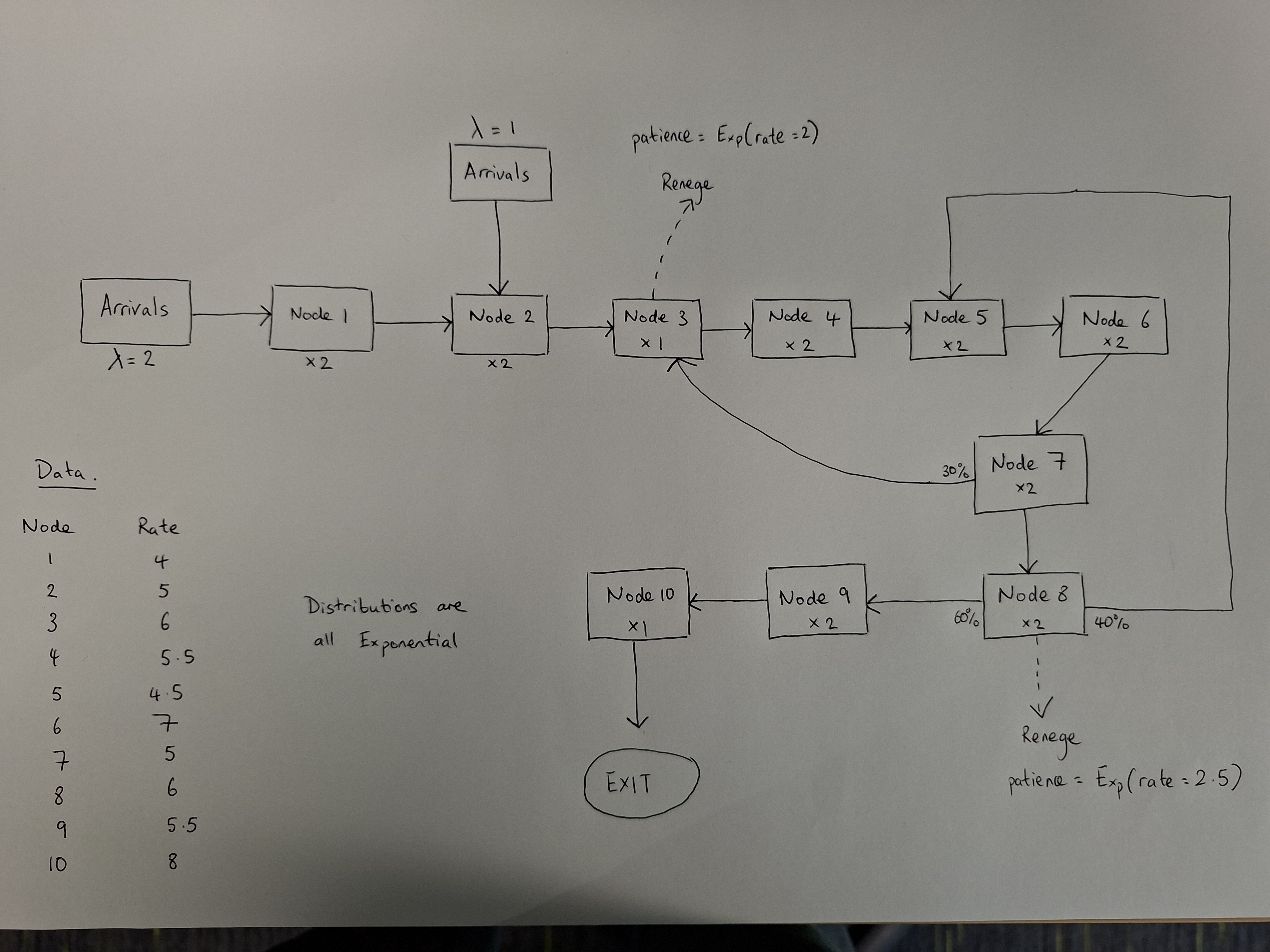}
    \caption{Case 8: Ten node network}
    \label{fig:case8_img}
\end{figure}

\end{appendices}

\end{document}